\documentclass[twoside,11pt]{article}

\usepackage[accepted]{melba}

\usepackage[utf8]{inputenc}
\usepackage{soul}
\usepackage{listings}
\usepackage{amsmath}
\usepackage{amsfonts}
\usepackage{amssymb}
\usepackage{graphicx}
\usepackage{caption}
\usepackage{subcaption,tikz}
\usepackage{algorithm2e}
\usepackage{stmaryrd}
\usepackage{bbm}
\usepackage{stmaryrd}
\usepackage{multicol, multirow}
\usepackage{stfloats}
\usepackage{scalerel}
\usepackage{hyperref}
\usetikzlibrary{positioning,fit}
\usepackage{float}
\usepackage{diagbox}
\DeclareMathOperator*{\logit}{logit}

\newcommand{\distTan}{{d}_{\text{Tan}}}
\newcommand{\distSog}{{d}_{\text{Sg}}}
\newcommand{\setS}{\mathcal{S}}
\newcommand{\setE}{\mathcal{E}_{\mathcal{S}}}

\newcommand{\lmsd}{\mathrm{LMSD}_{d}}
\newcommand{\lmsds}{\mathrm{LMSD}_{d^s}}

\newcommand{\dist}{\mathrm{dist}}

\newcommand{\crown}{C_{td}^\mathcal{N}}

\newcommand{\stp}{\mathrm{sTP}}
\newcommand{\stn}{\mathrm{sTN}}
\newcommand{\sfp}{\mathrm{sFP}}
\newcommand{\sfn}{\mathrm{sFN}}

\melbaheading{2023:013}{https://melba-journal.org/2023:013}{2023}{361-389}{01/2023}{09/2023}{Hamzaoui, Montagne, Renard-Penna, Ayache, Delingette}{MICCAI 2022 UNSURE Workshop}{Christian Baumgartner, Adrian Dalca, Koen Van Leemput, Raghav Mehta, Chen Qin, Carole Sudre, Ryutaro Tanno, William (Sandy) Wells}

\ShortHeadings{MACCHIatO}{Hamzaoui et al.}
\firstpageno{361}

\title{Morphologically-Aware Consensus Computation via Heuristics-based IterATive Optimization (MACCHIatO)}

\author{\name Dimitri Hamzaoui\orcid{0000-0003-2775-8594}\\ \addr Université Côte d’Azur, Inria, Epione Team, Sophia Antipolis, France \AND \name Sarah Montagne\\ \addr Academic Department of Radiology, Hôpital Pitié-Salpétrière, Assistance Publique des Hôpitaux de Paris, Paris, France\\ \addr GRC 5 Predictive Onco-Urology, Sorbonne University, Paris, France \AND \name Raphaële Renard-Penna\\ \addr Academic Department of Radiology, Hôpital Pitié-Salpétrière, Assistance Publique des Hôpitaux de Paris, Paris, France\\ \addr GRC 5 Predictive Onco-Urology, Sorbonne University, Paris, France \AND \name Nicholas Ayache\\ \addr Université Côte d’Azur, Inria, Epione Team, Sophia Antipolis, France \AND \name Hervé Delingette\orcid{0000-0001-6050-5949}\\ \addr Université Côte d’Azur, Inria, Epione Team, Sophia Antipolis, France}

\begin{document}

\maketitle

\begin{abstract}
The extraction of consensus segmentations from several binary or probabilistic masks is important to solve various tasks such as the analysis of inter-rater variability or the fusion of several neural network outputs. 
One of the most widely used methods to obtain such a consensus segmentation is the STAPLE algorithm. In this paper, we first demonstrate that the output of that algorithm is heavily impacted by the background size of images and the choice of the prior. We then propose a new method to construct a binary or a probabilistic consensus segmentation based on the Fr\'{e}chet means of carefully chosen distances which makes it totally independent of the image background size. We provide a heuristic approach to optimize this criterion such that a voxel's class is fully determined by {its voxel-wise distance to the different masks}, the connected component it belongs to and the group of raters who segmented it. 
We compared extensively our method on several datasets with the STAPLE method and the naive segmentation averaging method, showing that it leads to binary consensus masks of intermediate size between Majority Voting and STAPLE and to different posterior probabilities than Mask Averaging and STAPLE methods. Our code is available at \href{https~://gitlab.inria.fr/dhamzaou/jaccardmap}{https~://gitlab.inria.fr/dhamzaou/jaccardmap}.
\end{abstract}

\begin{keywords}
	Consensus, Distance, Heuristics, Optimization, STAPLE 
\end{keywords}

\section{Introduction}

The fusion of several segmentations into a single consensus segmentation is a classical problem in the field of medical image analysis related to the need to merge multiple segmentations provided by several clinicians into a single ``consensus'' segmentation. This problem has been recently revived by the development of deep learning and the multiplication of ensemble methods based on neural networks~\citep{nnunet}. One of the most well-known methods to obtain a consensus segmentation is the STAPLE algorithm~\citep{STAPLE}, where an Expectation-Maximization algorithm is used to jointly construct a consensus segmentation, and to estimate the raters' performances posed in terms of sensitivities and specificities. The seminal STAPLE method~\citep{STAPLE} creating a probabilistic consensus from a set of binary segmentations was followed by several follow-up works.
For instance, \cite{asman} replaced global indices of performance by spatially dependent performance fields and \cite{commowick} combined STAPLE with a sliding window approach to allow spatial variations of rater performances. Another improvement consisted in introducing the original image intensity information~\citep{asman_2}. 
Several alternatives to STAPLE were proposed, with a large diversity of approaches. Some of them decided to use a generative model but with different properties. For example, \cite{audelan} modeled raters' input maps by heavy-tailed distributions whose parameters are estimated by variational calculus, and \cite{5487420} presented a model using a random field learnt on the whole set to model the interaction between the intensity maps and the corresponding label maps. Methods based on deep learning were also conceived, as in \cite{neurips_nn} where two CNNs are trained together to estimate simultaneously the consensus segmentation and each rater's performance via an estimation of their spatial confusion matrices. Also in~\cite{arnaqueur} authors incorporate the expertise level of each rater and specific modules to better take into account disagreements between raters. However, those methods do not lead to explainable results and they require the collection of a preliminary training data on a consequent number of cases which make them not suitable on small datasets. In addition to those complex methods, several studies ~\citep{rohlfing,aljabar} show that simple majority voting (MV) could remain a suitable pick. However STAPLE and its simple yet robust probabilistic model remains the go-to method for consensus segmentation estimation~\citep{STAPLE,staple_use} despite suffering from several limitations, some of them already addressed in the literature~\citep{asman,commowick,asman_2} and some, to the best of our knowledge, never raised before.

In this article, we first analytically characterize the dependence of the STAPLE algorithm on the size of the background image and the choice of prior consensus probability. We then introduce an alternative consensus segmentation method, coined MACCHIatO, which is based on the minimization of the squared distance between each binary segmentation and the consensus. After choosing a distance between binary or probabilistic shapes, the consensus is thus posed as the estimation of the Fr\'{e}chet mean of this distance {(an extension of centroids to metric spaces)}, which is independent of the size of the background image for a well-chosen distance. We show that the adoption of specific heuristics based on morphological distances {(i.e. voxel-wise distances to the different binary masks based on morphological operations)} during the optimization allows to provide a novel binary or probabilistic globally consistent consensus method that creates masks of intermediate size between Majority Voting and the STAPLE methods.

This work extends our MICCAI-UNSURE 2022 paper \citep{mojito} by (1) Adding the Dice coefficient and its soft surrogates as distances between binary sets (2) Providing more mathematical details on baseline models and the STAPLE’s dependence on the background size and prior choice (3) Adding experiments and a dataset to justify the choice of selected heuristics and to analyze the impact on the consensus volume and computational time and (4) Expanding the discussion in various ways including detailing the limitations of the proposed approach.

\section{Estimation of a soft or hard consensus from binary segmentations}

In the remainder, we consider the problem of generating a consensus segmentation $T_n$, $1\leq n\leq N$ given $K$ binary segmentations $\setS = \{S^1, ..., S^K\}$, $S^k_n\in\{0,1\}$ of size $N$ provided by each rater $k$. The consensus segmentation may be either a \emph{hard} binary segmentation $T_n\in\{0,1\}$ or a \emph{soft} probabilistic segmentation $\tilde{T}_n\in[0,1]$, the tilde sign indicating that we are dealing with a continuous probabilistic consensus value, rather than a binary one. Given a soft consensus, one can easily generate a hard consensus by thresholding the soft consensus voxels at the 0.5 limit. Yet, this raises the issue of dealing with voxels that are exactly at the 0.5 value which can be either set arbitrarily to one of the 2 classes or set aside to a third class. 

In terms of probabilistic framework, the main approach is to consider that each observed binary segmentation $S^k$ results from a random process applied on a consensus segmentation $T$ which is captured by the likelihood distribution $p(S^k|T,\theta_k)$ also involving some parameters $\theta_k$ specific to each rater $k$. A prior probability on the consensus $p(T)$ is also defined related to the general \emph{a priori} knowledge about the consensus segmentation. Then a hard consensus can be obtained as a maximum likelihood $T=\arg \max_{M} p(\setS|M)$ or maximum \emph{a posteriori} estimate $U=\arg \max_{U} p(\setS|U)p(U)$ whereas a soft consensus is obtained as the posterior probability $p(\tilde{T}|\setS)=p(\setS|\tilde{T})p(\tilde{T})/p(\setS)$. The parameters $\theta_k$ are also estimated by maximum likelihood for hard consensuses or maximum marginal likelihood for soft ones.

We make use of the following notations~: $\mathrm{FP}_k$, $\mathrm{TP}_k$, $\mathrm{FN}_k$, and $\mathrm{TN}_k$ are respectively the number of false positives, true positives, false negatives, and true negatives between observed mask $S^k$ and consensus $T$, i.e. $FP_k=\sum_{n=1}^N S^k_n \wedge T_n$.

We consider as baseline methods to create a hard consensus the majority voting (MV) and the ML STAPLE {(Maximum Likelihood STAPLE, a binary version of STAPLE)} algorithms whereas mask averaging (MA) and STAPLE algorithm are baseline approaches for the soft consensus estimation. 
We describe below the hypotheses in terms of probability distribution associated with those baseline models and discuss their limitations.

\subsection{Majority Voting and Mask Averaging Models}
\label{subsec;mv-ma}

We first make the hypothesis of voxel independence, i.e. that the binary value of each voxel of an observed segmentation mask $S^k$ is independent of the values of other voxels~: $p(S^k|T)=\prod_{n=1}^N p(S^k_n|T_n)$. Furthermore, we consider that the prior and likelihood probability are simple Bernoulli distributions of the same parameter $b_n\in[0,1]$~: $p(S^k_n=1|b_n)=p(T_n=1|b_n)=b_n$. This means that the probability parameter $b_n$ is potentially different for all voxels, but the same for all raters~: $\theta_k=\theta=\{b_n\}$. Also, the observed masks $\setS$ do not directly depend on the consensus but share the same distribution. 

Therefore the likelihood of observing the whole segmentation data is then \[p(\setS|\theta)=\prod_{k=1}^K\prod_{n=1}^N b_n^{S^k_n}(1-b_n)^{1-S^k_n}=\prod_{n=1}^N b_n^{S_n^+}(1-b_n)^{S_n^-}\]
 where $S_n^+$ (resp. $S_n^-=K-S_n^+$) is the number of times voxel $n$ is equal to $1$ (resp. $0$) in the observed segmentation masks $S^k$,$1\leq k \leq K$. After maximizing the likelihood, one trivially gets the Bernoulli parameter as $p(S^k_n=1|b_n)=p(T_n=1|b_n)=\frac{S_n^+}{K}=b_n$, leading to the Mask Averaging consensus formula where the probability of having a foreground voxel is the frequency of positive voxels in the observed masks $S^k$. To estimate the hard consensus, one needs to maximize $p(T_n|b_n)$ thus leading to majority voting~: $T_n=1$ if $S_n^+>S_n^-$ and $T_n=0$ if $S_n^+<S_n^-$. 
 
 \paragraph{Limitations} Majority voting and mask averaging are simple and easy-to-understand mechanisms to choose a consensus. Yet they suffer from the fact that this decision is purely local without any influence from the neighboring pixels. This can lead to situations where the hard consensus includes some isolated voxels or has very irregular boundaries. {This is especially true for mask averaging, which does not have any mechanisms to enforce inter-rater consistency and that relies on the implicit assumption that the neighboring voxels of a segmented voxel are likely to be segmented, which is not the case on the boundaries}. Another limitation of majority voting is the case where the number of raters $K$ is even and therefore many decisions are ambiguous with as many foreground than background voxels. Finally, those simple models assume that all raters' contributions to the consensus are equal which may not be the case. In particular, an underperforming rater will bias the soft consensus with mask averaging.
 
\subsection{STAPLE model}

In the STAPLE algorithm~\citep{STAPLE}, all voxels are also assumed independent but the probability that $S_n^k$ is equal to $T_n$ depends on whether $T_n$ is a background or foreground voxel, and on the rater $k$. More precisely, $p(S^k_n=T_n|T_n=1)=p_k$ and $p(S^k_n=T_n|T_n=0)=q_k$ where $p_k$ is the sensitivity of rater $k$ and $q_k$ its specificity.

\paragraph{Prior Consensus}
The consensus prior probability is here supposed to factorize as the product of voxel priors $w_n$ values $p(T)=\prod_{n=1}^N P(T_n) = \prod_{n=1}^N w_n$. The original STAPLE paper~\citep{STAPLE} also introduced an Ising Markov random field model as a prior consensus probability to enforce that a voxel prior value depends on that of its neighbors. However, this approach leads to solving iteratively graph cuts problems and is not available in most widely used STAPLE implementations. Instead, the original paper assumes simple independent priors that lead to closed-form updates. Choosing $w_n=w=\frac{1}{2}$ is a non-informative prior but another common choice is to have a spatially uniform value $w_n=w = \frac{1}{NK}\sum_{n,k} S_n^k$ which is the average relative size of the foreground object in the observed segmentation masks. 
We further consider more general priors of the form $w=\frac{A}{N^\alpha}$, with $A$ a constant independent of the image size, and $\alpha\in\mathbb{N}$ an exponent. The non-informative case $w_n=0.5$ corresponds to $\alpha=0$ while the average object size to $\alpha=1$.
 
\paragraph{Maximum likelihood STAPLE (ML STAPLE)} The likelihood of the observed data simply writes as $\mathcal{L}(T,
\theta)=\prod_{k=1}^K p_k^{\mathrm{TP}_k}(1-p_k)^{\mathrm{FN}_k}q_k^{\mathrm{TN}_k}(1-q_k)^{\mathrm{FP}_k}$ and does not involve the prior on the consensus. There is no closed-form expression for the estimation of the rater parameters ($p_k, q_k$) and the hard consensus ($T$) maximizing the likelihood. But an iterative maximization of the likelihood is possible by setting its derivatives to zero which leads to the update equation~: 

\begin{align}
  p_k=\frac{\mathrm{TP}_k}{\mathrm{TP}_k+\mathrm{FN}_k} & & q_k=\frac{\mathrm{TN}_k}{\mathrm{TN}_k+\mathrm{FP}_k} \label{eq;stapleML1} \\
  s_n^+=\prod_{k=1}^K p_k^{S^k_n}(1-p_k)^{1-S^k_n} && s_n^-=\prod_{k=1}^K q_k^{1-S^k_n}(1-q_k)^{S^k_n} \label{eq;stapleML2}\\
  T_n=1~\mathrm{if}~ s_n^+> s_n^-&&T_n=0~\mathrm{if}~ s_n^+< s_n^- \nonumber
\end{align}

\paragraph{Maximum marginal likelihood (MML STAPLE)}
 The \emph{marginal likelihood} or \emph{evidence} writes as $p(\setS|\theta)= \prod_{n=1}^N (w_n \prod_k p_k^{S_n^k} (1-p_k)^{1-S_n^k} + (1-w_n) \prod_k q_k^{1-S_n^k} (1-q_k)^{S_n^k})$ and is only a function of the rater parameters $\theta_k$. Its maximization is not tractable in closed form but the expectation-maximization algorithm provides a way to estimate some local maxima. The E-step consists in evaluating the posterior probability from Bayes law with the current estimated sensitivities and specificities~:
\begin{equation}\label{eq;posterior}u_n = p(\tilde{T}|\theta,\setS)=\frac{w_n \prod_k p_k^{S_n^k} (1-p_k)^{1-S_n^k}}{w_n \prod_k p_k^{S_n^k} (1-p_k)^{1-S_n^k} + (1-w_n) \prod_k q_k^{1-S_n^k} (1-q_k)^{S_n^k}} \end{equation}. 


The M-step updates the parameters $p_k$ and $q_k$ as follows~:
\begin{align}
  p_k=\frac{\sum_{n, S_n^k=1} u_n}{\sum_n u_n}=\frac{\stp_k}{\sfn_k + \stp_k} & & q_k = \frac{\sum_{n, S_n^k=0} (1-u_n)}{\sum_n (1-u_n)} = \frac{\stn_k}{\stn_k + \sfp_k}
\end{align}
where $\stp_k$, $\stn_k$, $\sfp_k$, $\sfn_k$ are the "soft extension" of the number of true positive, true negative, false positive, and false negative voxels from rater $k$. 

\subsubsection{Influence of the prior term} We can better understand the influence of the prior when estimating the probability to belong to a consensus by writing its logit $\logit(u_n) = \ln{(\frac{u_n}{1-u_n})}$ from Eq.\ref{eq;posterior}~:
\begin{align}
\label{eq;logitun}
 \logit{(u_n)} &= \logit(w_n) + \sum_{k, {S_n^k}=1} \log \left( \frac{p_k}{1-q_k} \right) +
  \sum_{k, {S_n^k}=0} \log\left(\frac{1-p_k}{q_k}\right)\end{align} 
Thus, we see that to estimate $u_n$ each foreground voxel of rater $k$ "votes" with a (usually) positive quantity $\log \left( \frac{p_k}{1-q_k} \right)$ whereas each background voxel "votes" with a (usually) negative quantity $\log\left(\frac{1-p_k}{q_k}\right)$. Then the prior term $\logit(w_n)$ biases this vote depending on whether $w_n$ is greater or smaller than $\frac{1}{2}$.

\subsubsection{Influence of the background size}
\label{subsec;backgroundSizeInfluence}
In many cases, the size $N$ of images that contain the objects delineated by the raters is arbitrary since it can be the size of the original image (with a large value of $N$) or the size of a restricted region of interest (with a small value of $N$). It is therefore important to estimate the influence of the background size, i.e. the number of true negative voxels $\mathrm{TN}_k$, in the estimation of the hard and soft consensus. 

\paragraph{Influence on hard consensus} Based on Eqs.\ref{eq;stapleML1} and \ref{eq;stapleML2}, the sensitivity and coefficient $s_n^+$ are not influenced by $\mathrm{TN}_k$, but the specificities are. More precisely, we have $q_k=1-\frac{\mathrm{FP}_k}{\mathrm{TN}_k} +O((\mathrm{TN}_k)^{-2})$, and therefore the quantity $s_n^-$ tends towards $0$ when $\mathrm{TN}_k$ reaches large values. This implies that the hard consensus converges towards the union of all observed segmentation masks when the background size becomes large.
 
\paragraph{Influence on soft consensus} The posterior probability $u_n$ and specificities $q_k$ are mainly impacted by the increase of the background size, while the sensitivities are more marginally influenced. The nature of the soft consensus depends on the $\alpha$ exponent of the prior expression $w_n=\frac{A}{N^\alpha}$, and in particular we have~:

\begin{align*}
\logit{(u_n)}= (\sum_{k=1}^K S_n^k - \alpha)\log N +\log A + \ln{(\frac{p_k}{\sfp_k})} + \sum_{k, S_n^k=0} \ln{(1-p_k)} +O(N^{-2})
\end{align*}

\begin{figure}
    \begin{subfigure}{0.48\textwidth}
        \centering
        \includegraphics[width = 0.95\textwidth]{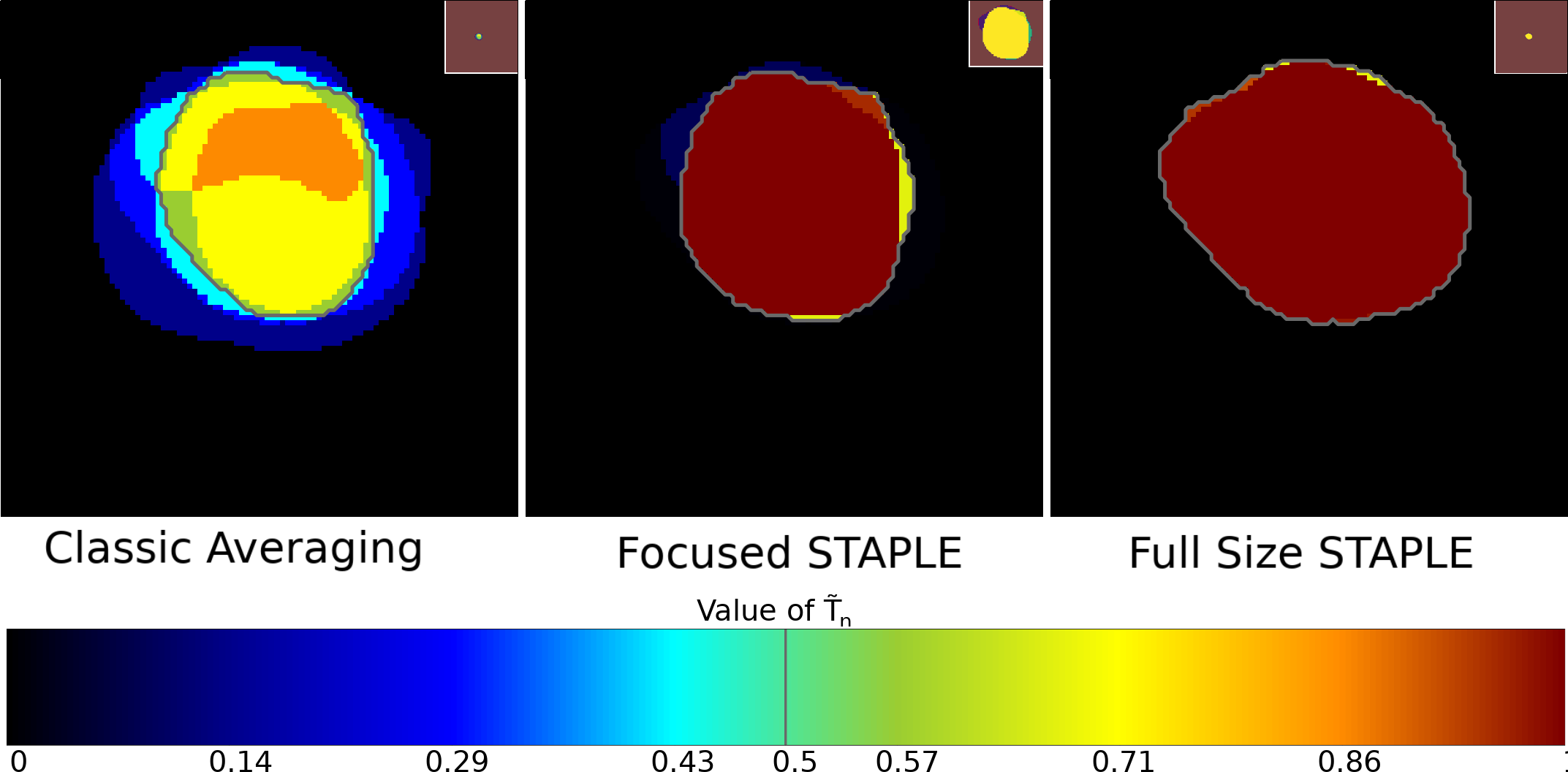}
        \caption{Impact of background size on a simple 2D case with 7 raters (the 7th is an empty map), with STAPLE computed on a 67$\times$61 image (middle), and 640$\times$640 (right) with $w = (\sum_{n,k} S_n^k) / NK$. The relative size of the structure can be seen at the top right corner.}
        \label{fig:1a}
      \end{subfigure}
      \hfill
      \begin{subfigure}{0.48\textwidth}
        \centering
        \includegraphics[width = 1\textwidth]{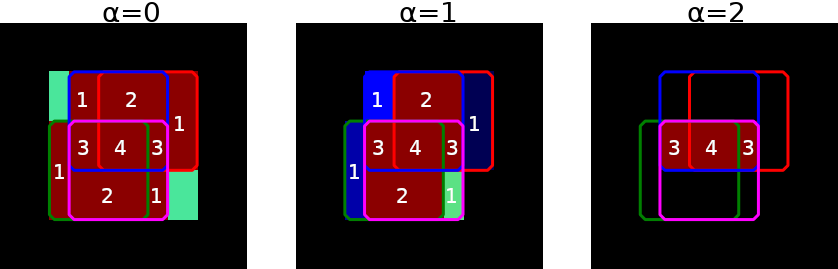}
        \caption{Limits of STAPLE algorithm for 3 different values of $\alpha$ on a toy example with 4 raters providing the red, blue, green and magenta contours. The figures are the number of raters who segmented this zone, and the colors are the probability of the soft consensus with colormap borrowed from Fig.\ref{fig:1a}.}
        \label{fig:1b}
      \end{subfigure}
      \caption{Impact of STAPLE hyperparameters and background size on the soft consensus}
 \end{figure}

{A direct consequence of this formula is that the background size impacts the obtained consensus, as can be seen in Fig.~{\ref{fig:1a}} where the consensus obtained when applying STAPLE on a bounding box tightly surrounding the organ (referred to as \emph{Focused STAPLE} in Fig.~{\ref{fig:1a}}) appears as smaller and with more non-binary values than the one computed on the whole image (referred to as \emph{Full size STAPLE} in Fig.~{\ref{fig:1a}}). Comparisons between STAPLE computed on both volumes are available in Tab.~{\ref{tab:9}} in the appendices.}
Moreover, as seen in Fig.\ref{fig:1b}, the soft consensus when having a large background size depends on the value of $\alpha$, with larger $\alpha$ corresponding to smaller consensuses. The detailed proof is presented in Appendix~\ref{STAPLE_back}.

\paragraph{Removing the Influence of the background size} We explore under which conditions the STAPLE model leads to consensus estimations that are independent of the background size. A first simplification of the model is to assume that all raters perform equally $p_k=p$, $q_k=q$. In this case, the global specificity maximizing the likelihood is $q=\frac{\sum_{k=1}^K \mathrm{TN}_k}{\sum_{k=1}^K \mathrm{TN}_k+\mathrm{FP}_k}$ which is still dependent on the size of the background through $\mathrm{TN}_k$. 

A second simplification is to consider that each rater sensitivity and specificity are equal, i.e. $p_k=q_k=\gamma_k$. This implies that the rater performance is independent of the fact the consensus voxel is in the background or foreground. In this case, the parameter $p_k=q_k=\gamma_k$ can be interpreted as the accuracy parameter and its optimization leads to $\gamma_k=\frac{\mathrm{TP}_k+\mathrm{TN}_k}{N}$. It is easy to see that in that case, $\frac{s_n^+}{s_n^-}=\left(\frac{\gamma_k}{1-\gamma_k}\right)^{S_n^+ - S_n^-}$, and therefore the maximum likelihood is equivalent to majority voting when $\gamma_k>\frac{1}{2}$ which is independent of background size. With this simplification, and from Eq.\ref{eq;logitun}, the soft consensus obtained by maximizing the marginal likelihood with a non-informative prior $w_n=\frac{1}{2}$, is such that $\logit(u_n)=(S_n^+ - S_n^-)\logit(\gamma_k)$. The value of $\gamma_k$ depends on the background size, but whether a voxel is more likely to be a background pixel $u_n>\frac{1}{2}$ does not depend on the background size. 

\subsubsection{Limitations}
The STAPLE algorithm addresses the problem of taking into account the performance of raters when building a consensus segmentation. However, this approach has the drawback of being dependent on the choice of the prior, and the background size.  This dependence of the STAPLE consensus can be explained by the fact that it is a generative model which should explain the foreground and the background voxels separately. When assuming that the rater performance is the same in both background and foreground, then the model becomes equivalent to majority voting. {This dependence is a subject of concern as STAPLE is often used as a standard in label fusion works. To improve the robustness of comparisons with novel methods and decrease the impact of this hidden hyperparameter, researchers may compute STAPLE consensus using several bounding boxes, or at least indicates the size of the bounding box on which STAPLE was applied.}\\
The use of local sliding windows in STAPLE as in \cite{commowick} can somewhat mitigate the background size effect, but smaller structures in images can still be impacted and the window size remains a hyperparameter which is difficult to set.\\

\section{MACCHIatO framework}

\subsection{Main approach description}

In the previous section, we have seen that only the majority voting and mask averaging algorithms lead to a consensus that is independent of the background size. Yet, those algorithms are purely local at the voxel level and can lead to irregular boundaries or isolated voxels. 

In this section we introduce a new framework to compute soft and hard consensuses that are i) invariant from the background size and ii) dependent on the global morphology of each binary object. This approach is coined MACCHIatO for Morphologically-Aware Consensus Computation via Heuristics-based IterATive Optimization. 

\paragraph{Distance-based approach} We formulate the estimation of a hard consensus $T$ as the minimization of the sum of the square distance between the consensus $T$ and each observed binary mask $S^k$~:
\begin{equation}
\label{eq;variance-hard}
T = \arg \min_{M\in\{0,1\}^N} \sum_{k=1}^K d(M, S^k)^2
\end{equation}
 where $d(T,S^k)$ is a distance as defined in \cite{deza} between the two masks $S^k$ and $T$. This is equivalent to estimating the consensus as a maximum likelihood where the likelihood can be written as $p(S^k|T)\propto \exp(-\lambda d(T, S^k)^2 )$.  {We can note that the squared sum $\sum_{k=1}^K d(M, S^k)^2$ also corresponds to the definition of a Fr\'{e}chet variance. Based on this interpretation, $T$ appears as the Fr\'{e}chet mean of $\setS$ i.e. its centroid in the metric space defined by $d$.} 
 
 \paragraph{Link with baseline models} In section~\ref{subsec;backgroundSizeInfluence}, we have seen that when the sensitivity and specificity are equal, the maximization of the STAPLE model leads to the majority voting algorithm. In this case, we can write the likelihood $p(S^k|T)=\gamma_k^{\mathrm{TP}_k+\mathrm{TN}_k}(1-\gamma_k)^{\mathrm{FP}_k+\mathrm{FN}_k}$ (where $\gamma_k$ is the accuracy parameter) which is a product of $N$ independent Bernoulli distributions. Since the Bernoulli distribution is a member of the exponential family~\citep{exponentialFamily}, it can be also written as $p(S^k|T)\propto \exp(-\lambda_k (\mathrm{FP}_k+\mathrm{FN}_k))$ where $\lambda_k=\logit(\gamma_k)$. The number of false positives or false negatives $\mathrm{FP}_k+\mathrm{FN}_k$ is the number of elements of symmetric difference between the two sets $S^k$ and $T$~: $\mathrm{FP}_k+\mathrm{FN}_k=|T\Delta S^k|=|(T\cup S^k)\setminus(T\cap S^k)|$ and is also called the \emph{Hamming distance} in information theory. Thus, by choosing $d(T, S^k)=\sqrt{|T\Delta S^k|}$, the maximum likelihood leads to majority voting consensus (as detailed in Appendix~\ref{MV_proof}).

\paragraph{Soft consensus framework} On the baseline models, soft consensuses were obtained as posterior probabilities of having a consensus from the observed binary masks. However, from the likelihoods $p(S^k|\tilde{T})\propto \exp(-\lambda d(\tilde{T}, S^k)^2 )$, the computation of the posterior $p(\tilde{T}|\setS)$ may not be tractable due to the difficulty of computing the normalization constant. Instead, we propose to approximate $p(\tilde{T}_n|\setS)$ by the quantity $\tilde{U}_n\in[0,1]$ such that $\tilde{U}\in [0,1]^N$ minimizes the quantity~:
\begin{equation}
\label{eq;variance-soft} \tilde{U} = \arg \min_{\tilde{X}\in [0,1]^N} \sum_{k=1}^K d^s(\tilde{X}, S^k)^2 \end{equation}
where $d^s( \tilde{X}, S^k)$ is a distance between the probabilistic array $\tilde{X}$ and the binary mask $S^k$. More precisely, the distances $d^s(\tilde{X}, S^k)$ considered are \emph{soft surrogate} of the distance between binary sets $d(\tilde{X} ,S^k)$ such that $d^s(\tilde{X}, S^k)^2=d(\tilde{X}, S^k)^2$ when $\tilde{X}\in\{0,1\}^N$. For instance, the distance $d(\tilde{X},S^k)=\|\tilde{X}-S^k\|$ is a soft surrogate of the Hamming distance since $|\tilde{X}\Delta S^k|=\|\tilde{X}-S^k\|^2$. Besides it is clear that the mask averaging (MA) method is a soft consensus minimizing the following squared sum $\sum_{k=1}^K \|\tilde{U}-S^k\|^2$.
 
\paragraph{Optimization approach} The estimation of the soft and hard consensus is independent of the background size if the distance $d(T,S^k)$ is invariant to the number of true negatives. Besides, unlike the MV and MA algorithms, the optimization cannot be performed at the voxel level when the distance cannot be split voxelwise. Instead of optimizing the whole foreground object, we chose to consider each connected component separately from each other to obtain more coherent results. Finally, we further split the optimization into subcrowns with various heuristics to speed up the computation.

\subsection{Distances between binary masks} We detail below the selected distances between binary sets that are considered and their associated soft surrogates. 
We mainly focus on distances based on two widely used methods to measure the overlap between binary segmentations~: the Jaccard and Dice coefficients.

\paragraph{Jaccard distance} The Jaccard coefficient (aka IoU) between binary masks $A$ and $B \in \{0, 1\}^N$ is defined as~: $\text{Jac}(A,B)=\frac{|A\cap B|}{|A\cup B|}$. In~\cite{kosub}, it is shown that its complementary to 1 $\dist_J (A, B) = 1 - \text{Jac}(A, B) = \frac{|A\Delta B|}{|A\cup B|}$ is a metric between binary sets following the triangular inequality. Several formulations of soft surrogates exist that extend the Jaccard distance. We focused specifically on two of them~: the Soergel metric~\citep{spath,deza} $\distSog(x, y) = \frac{\sum_i \max(x_i, y_i) - \min(x_i, y_i)}{\sum_i \max(x_i, y_i)}$ which follows the triangular inequality but is not differentiable, and the widely-used Tanimoto distance~\citep{willett,deza,leach2007} $\distTan(x, y) = 1 - \frac{\sum_i x_i y_i}{\sum_i x_i^2+y_i^2-x_i y_i} = \frac{||x-y||^2}{||x-y||^2+<x, y>}$.

\paragraph{Dice coefficient} 
It is defined as $\text{DSC}(A, B) = \frac{2|A\cap B|}{|A| + |B|}$ and is widely used in image segmentation as a performance index. Indeed, the Dice index is equal to the F1-score and corresponds to the harmonic mean of the sensitivity and positive predictive value. It is closely related to the Jaccard coefficient as $\text{DSC}(A, B) = \frac{2 \text{Jac}(A,B)}{1 + \text{Jac}(A,B)}$. The Dice distance $\dist_D (A, B) = 1 - \text{DSC}(A, B)$ is a near-metric i.e. it respects a relaxed form of the triangular inequality~\citep{diceinequality}. Soft surrogates of the Dice distance have been developed especially as a loss function in deep learning. We consider in the remainder two main extensions of the Dice distance~\citep{junma} on non-binary sets defined as ${d}_{\text{pSD}} (x, y) = 1 - \frac{2\sum_i x_i y_i}{\sum_i x_i^p + \sum_i y_i^p}$ where $p \in \{1, 2\}$.\\\newline

By construction, all those distances only depend on segmented pixels and are independent of the background size. Note that both distances are extended to get a null distance between two empty sets. {Using those distances in the Fr\'{e}chet variance computation, the inclusion of voxels segmented by a large number of raters (resp. a few raters) decreases (resp. increases) its value.} The different formulations of the MACCHIatO framework are summarized in table~\ref{tab:dist}.

\begin{table}[!htbp]
  \centering
  \caption{Distances between binary sets and their soft surrogate considered to compute hard and soft consensuses with the MACCHIatO framework}
  \resizebox{\textwidth}{!}{
  \begin{tabular}{|c|c|c|c|c|}
    \hline 
    
    \multirow{2}{*}{\shortstack{Hard Consensus\\Method}} & \multirow{2}{*}{\shortstack{Soft Consensus\\Method}} & \multirow{2}{*}{\shortstack{Distance}} & \multirow{2}{*}{\shortstack{Soft Surrogate}} & \multirow{2}{*}{\shortstack{Computation-level}} \\ 
    \ & \ & \ & \ & \ \\\hline
    Majority Voting & Mask Averaging & $|A\Delta B|$  & $\|x-y\|$ & Voxel-level\\  \hline
    ML STAPLE & MML STAPLE & NA & NA & Image-level \\  \hline
    \multirow{2}{*}{\shortstack{MACCHIatO-J}} &  MACCHIatO-TJ &  \multirow{2}{*}{\shortstack{Jaccard $d_J$}} & Tanimoto  $\distTan$  & \multirow{4}{*}{\shortstack{Connected\\component\\level}} \\\cline{2-2} \cline{4-4}
    \ &  MACCHIatO-SJ & \  & Soergel  $\distSog$ & \ \\  \cline{1-4}
    \multirow{2}{*}{\shortstack{MACCHIatO-D}} &  MACCHIatO-1SD &  \multirow{2}{*}{\shortstack{Dice $d_D$}} &  ${d}_{\text{1SD}}$ & \ \\  \cline{2-2} \cline{4-4}
    \ &  MACCHIatO-2SD & \ &  ${d}_{\text{2SD}}$ & \ \\  \hline
  \end{tabular}}

  \label{tab:dist}
\end{table}

\subsection{Heuristic computation based on morphological distance and crowns} \label{subcrown_def}

\paragraph{Domain of optimization} Since the distances listed in the previous section are independent of the number of true negatives, their computations can be restricted to the union of all rater masks~: $\setE = \{n|\sum_{k=1}^K S_n^k>0\}$. Furthermore, we consider that to decide whether a voxel belongs to the consensus, one should only take into account the regional context associated with the connected components surrounding that voxel, since far-away components may not be relevant. 
Therefore, we choose to minimize separately the Fr\'{e}chet variances of Eqs.~\ref{eq;variance-hard} and \ref{eq;variance-soft} for each connected component $St$ of the masks union $\setE$. 
Therefore, in practice, we minimize the \emph{Local Mean Squared Distance} between $\setS$ and the consensus~: $\mathrm{LMSD}_d (\setS, M) = \sum_{{St}\subset\setE} \frac{1}{K} \sum_k d(S^k_{\|St} ,M_{\|St})^2$ where $S^k_{\|St}$ (resp. $M_{\|St}$) are the restriction of the binary masks $S^k$ (resp. $M$) to the connected component $St$. {A benefit of this choice is that the determination of the Fr\'{e}chet Mean behaves similarly to a structure-wise MV, as the Fr\'{e}chet Mean of components segmented by less than half of the raters is the null set. However, contrary to MV, raters who do not segment a component kept by the majority of raters do not bias its consensus segmentation, as their contribution to the associated LMSD is {$\frac{1}{K}\delta_\emptyset (M_{\|St})$} = 0 and does not impact the Fr\'{e}chet mean.}
To lighten notations, we drop the $St$ index in the remainder. It is equivalent to considering that $\setE$ has only one single connected component. 

\begin{figure}[!htbp]
\centering
\begin{subfigure}{1.0\linewidth}
  \centering
  \includegraphics[width=1.0\linewidth]{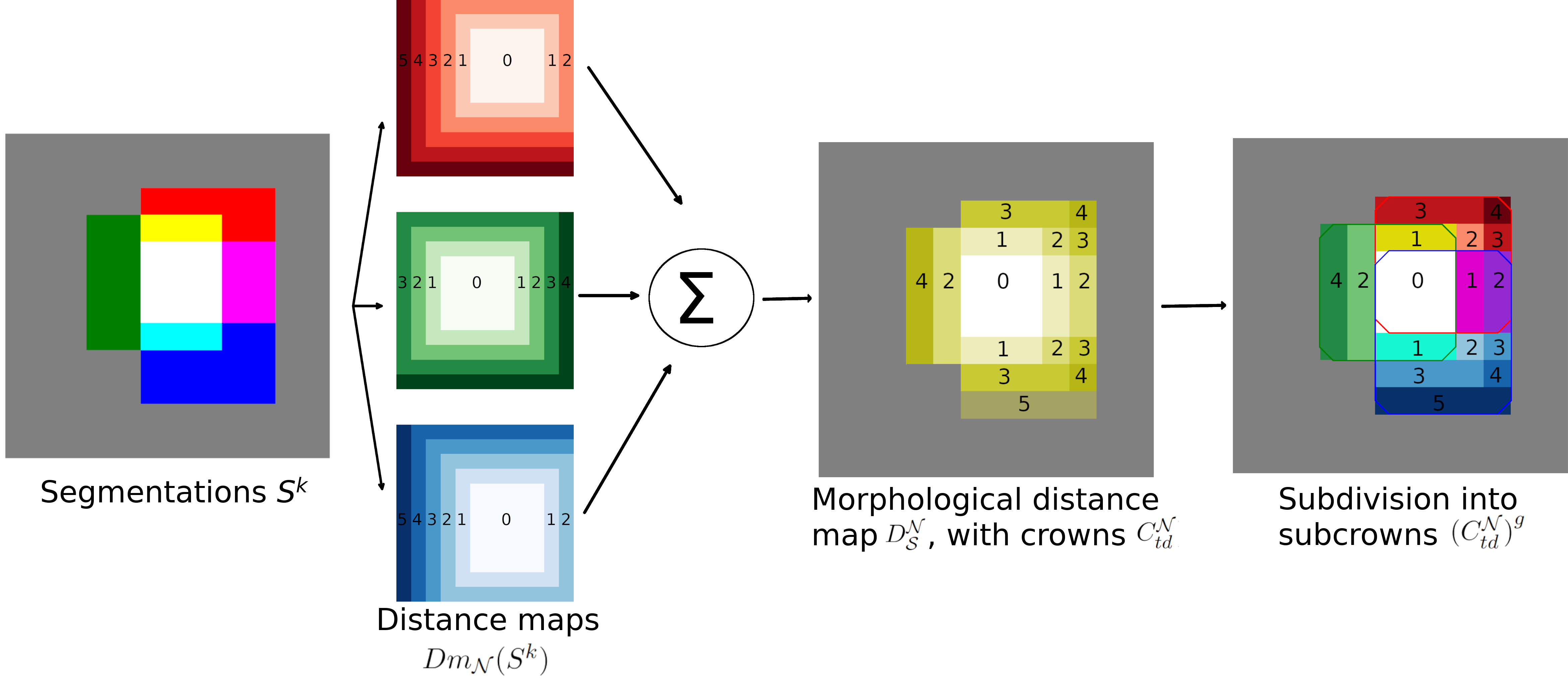}
  \subcaption{}
  \label{fig:2a}
\end{subfigure}
\vfill
\begin{subfigure}{0.35\linewidth}
\centering
  \includegraphics[width=1.0\textwidth]{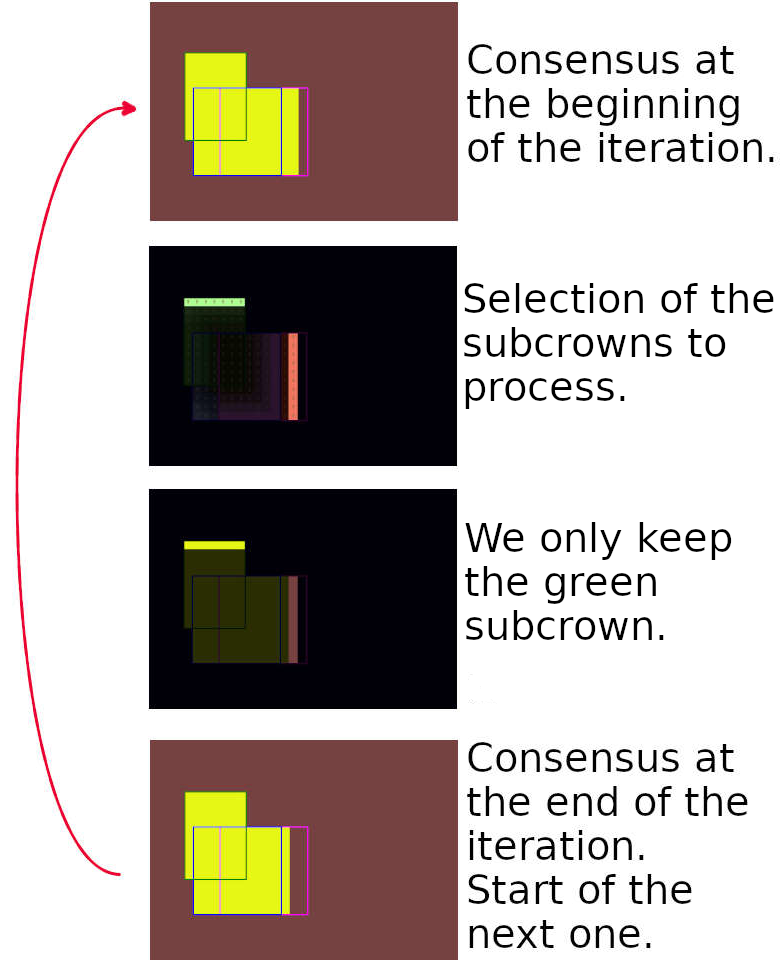}
  \subcaption{}
\label{fig:2b}
\end{subfigure}
\hfill
\begin{subfigure}{0.6\linewidth}
\centering
  \includegraphics[width=1.0\textwidth]{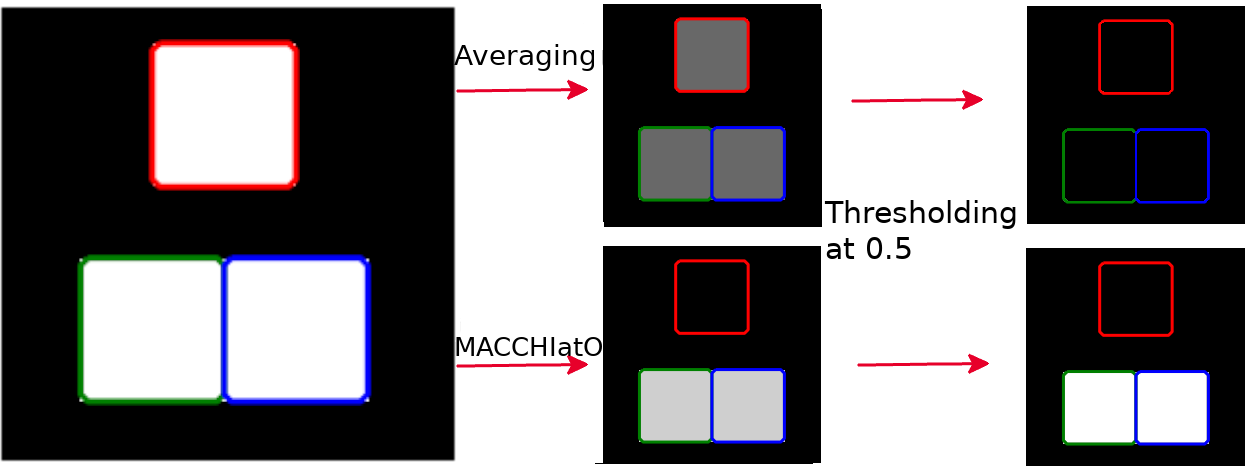}
  \subcaption{}
\label{fig:2c}
\end{subfigure}
\caption{{(a) Construction of the heuristics. From left to right: Segmentations by 3 raters (red, green, and blue); computation of the associated distance maps $Dm_{\mathcal{N}}(S^k)$; merging into the morphological distance map $D_\setS^\mathcal{N}$ restricted to the voxels segmented at least one; subdivision into subcrowns (1 color = 1 subcrown) based on morphological distance and raters). (b) An iteration of the shrinking approach with the selection of sub-crowns and the evaluation of their contribution to the $\lmsd$
(c) Application of mask averaging and soft MACCHIatO on a toy example with three segmentations (red, green, and blue contours). After thresholding, averaging gives an empty segmentation whereas the soft MACCHIatO method is more inclusive and outputs one connected component.}}
\end{figure}

\paragraph{Subcrown-based optimization}
The minimization of the Fr\'{e}chet variance is a combinatorial problem with a complexity of $2^{|\setE|}$ for the naive approach. Furthermore, it may lead to several global minima when the number of raters $K$ is small. For those reasons, we propose instead to seek a local minimum of the Fr\'{e}chet variance by introducing some heuristics in the optimization. With this approach, the local minimum has a lower complexity to compute and, by construction, is maximally connected to avoid isolated voxels.\\\newline
More precisely, instead of a computationally expensive per voxel minimization of the Fr\'{e}chet variance, we decompose the set $\setE$ into a set of {\em subcrowns} that take into account the global morphological relationships between each rater mask. The formal definition of subcrowns requires the specification of distance maps $Dm_{\mathcal{N}}(S^k)$ to each binary mask $S^k$ on $\setE$ according to a chosen neighborhood $\mathcal{N}$. This one can be
 either the 4 or 8 (resp. 6 or 26) connectivity in 2D (resp. 3D).  The distance $Dm_{\mathcal{N}}(S^k)$ is set to 0 for all voxels inside the object $S^k$.\\\newline
 The global morphological distance map is the sum of those distance maps \[D_\setS^\mathcal{N} = \sum_{S^k \in \setS}Dm_{\mathcal{N}}(S^k)\] for all raters on $\setE$. A \emph{crown} $\crown$ is defined as the set of voxels having the same global morphological distance $td$. Those crowns realize a partition of $\setE$ ($\setE = \coprod_{td} \crown$), and the $0$-crown corresponds by construction to the intersection of all masks in $\setS$.\\\newline We further split each crown as a set of \emph{subcrowns} by grouping the voxels that have been produced by the same set of raters. In other words, a subcrown corresponds to a set of voxels located at the same morphological distance from the intersection of all rater masks and which have 
been segmented by the same group of raters, as seen in Fig.~\ref{fig:2a}. Formally, a subcrown is noted ${(\crown)}^g$ where the superscript $g$ corresponds to a group of raters and subcrowns realize a partition of a crown~:

\begin{align}
\label{decomposition}
\crown = \coprod_{g\in \mathcal{P}(\llbracket 1, K \rrbracket)} {(\crown)}^g, \text{with } ({\crown})^g = \{n |n\in\crown\ \&\ \forall k\ S_n^k = (k \in g)\}
\end{align} where $\mathcal{P}(\llbracket 1, K \rrbracket)$ is the power set (i.e. the set of all subsets) of the first K integers.\\\newline
{The process for the construction of subcrowns is illustrated in Fig.~{\ref{fig:2a}}}

\subsection{Hard consensus algorithm}
\label{MACCHIatO}

The optimization proceeds in a greedy fashion by iteratively removing or adding subcrowns to the current estimate of the consensus until the $\lmsd$ criterion stops decreasing. In Alg.~\ref{alg:hard}, we use two concurrent strategies~: either we start from the union of all masks and then remove subcrowns with decreasing distances (a straegy illustrated in Fig.~\ref{fig:2b}, or we start with the crown with the minimum distance and then add subcrowns of increasing distances. {Both growing and shrinking strategies are applied as the greedy process can lead to different results, and we keep the consensus associated with the minimum $\lmsd$ of both strategies and the null set. The latter is also tested in the last stage since the distance of a set $M$ to the null set is $\delta_\emptyset (M)$, for both Dice and Jaccard distances. This discontinuity is not compatible with the iterative process and calls for a independent test.} 

\begin{figure}[!htbp]
    \centering
    \includegraphics[width = 1.0\textwidth]{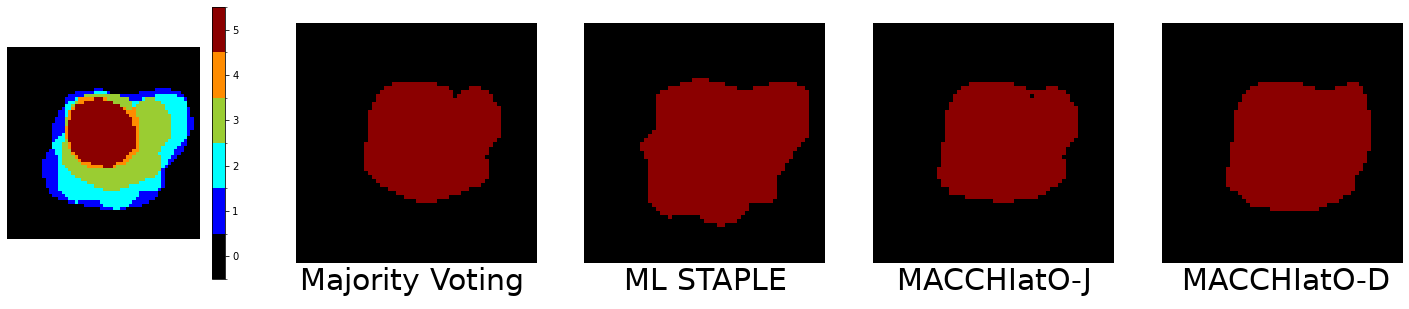}
    \caption{Comparison of several hard consensus methods on a 2D slice with 5 raters using MV, ML STAPLE and both hard MACCHIatO. On the left is indicated the number of raters who segmented each pixel.} 
    \label{fig:3}
  \end{figure}

Examples of consensuses obtained with this strategy can be seen in Fig.~\ref{fig:3}. Thus, the resulting consensus leads to a consistent grouping since all voxels belonging to the same connected component, having the same morphological distance, and being generated by the same group of raters will end up in the same class. Alternative optimization approaches could have been based on adding or removing single voxels (smaller than subcrowns) or crowns (larger than subcrowns). While voxel-based minimization would be very time-consuming, especially in 3D, conversely crown-based would lead to suboptimal results as crowns can be fairly large. 
Thus, the Morphologically-Aware Consensus Computation via Heuristics-based IterATive Optimization (MACCHIatO) algorithm is designed to be a good compromise between computational efficiency and consistency, with a number of iterations exponentially depending on $K$ but which is lower than the naive $2^{|\setE|}$ complexity.


\begin{algorithm}[!htbp]

\DontPrintSemicolon
\caption{Hard consensus algorithm.}
\label{alg:hard}
\SetKwInput{Initialization}{Initialization}
\KwIn{$\setS\ \text{segmentation maps},\ \mathcal{N}\ \text{neighborhood},\ {d}\ \text{distance}$}
\KwResult{$T$}
\Initialization{Computation of $D_\setS^\mathcal{N},\ td_u = \max(D_\setS^\mathcal{N})$, $td_i = \min(D_\setS^\mathcal{N})$;\\ $T^u = \bigcup_{k} S^k$; $T^i = \{n | (D_\setS^\mathcal{N})_n=td_i \}$}
\While(\tcp*[h]{Shrinking strategy}){$\lmsd(T^u, \setS)\text{ decreases}$}
{
   
    \For{$g \in \mathcal{P}(\llbracket 1, K \rrbracket)$}
    {\If{$\lmsd((T^u/(C_{td_u}^\mathcal{N})^g), \setS) < \lmsd(T^u, \setS)$}
    {$T^u \gets T^u/{(C_{td_u}^\mathcal{N})^g}$}}
    $td_u \gets \max(\{x \in D_\setS^\mathcal{N}| x< td_u \})$
}

\While(\tcp*[h]{Growing strategy}){$\lmsd(T^i, \setS)\text{ decreases}$} 
{
     \For{$g \in \mathcal{P}(\llbracket 1, K \rrbracket)$}
    {\If{$\lmsd((T^i \cup (C_{td_i}^\mathcal{N})^g), \setS) < \lmsd(T^i, \setS)$}
    {$T^i \gets T^i \cup (C_{td_i}^\mathcal{N})^g$}}
    $td_i \gets \min(\{x \in D_\setS^\mathcal{N}| x > td_i\})$
}
$T \gets \arg \min\limits_{T\in\{T^u, T^i,\emptyset\}} \lmsd(T, \setS)$
\end{algorithm}

\subsection{Soft consensus algorithm}

The estimation of a probabilistic or soft consensus is based on the minimization of the sum of square surrogate distances as displayed in Eq.~\ref{eq;variance-soft} and the optimization is split for each connected component of the mask union $\setE$.

\begin{figure}[!htbp]
    \centering
    \includegraphics[width = 0.6\textwidth]{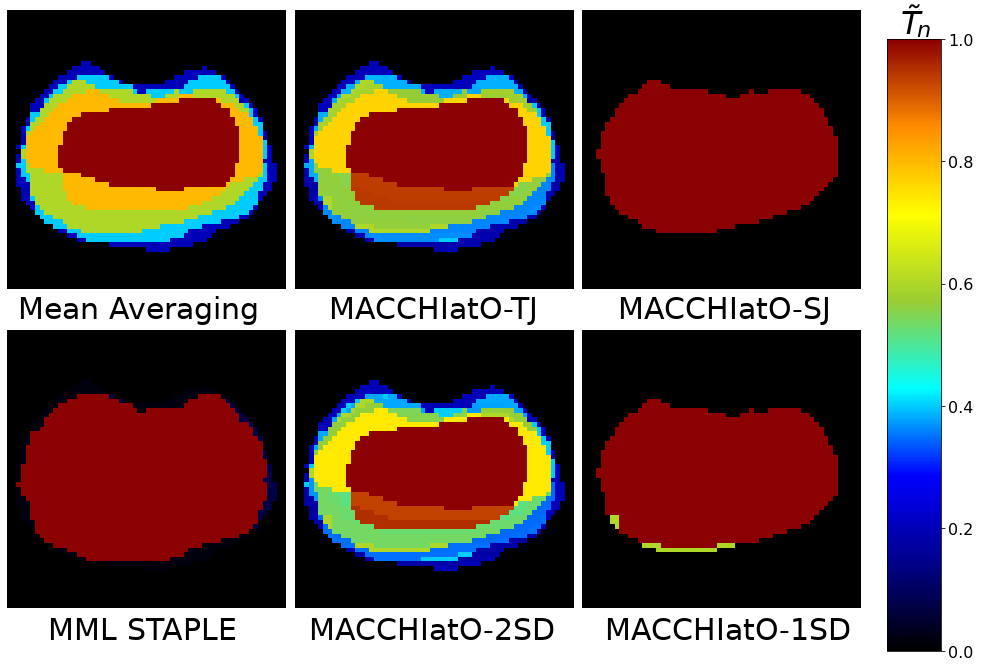}
    \caption{Comparison of several soft consensus methods on a 2D case with 5 raters using MA, STAPLE and MACCHIatO with different distances.} 
    \label{fig:4}
  \end{figure}

The \emph{soft MACCHIatO} algorithm extends the previous approach to minimize the criterion $\lmsds(\tilde{T}, \setS)$. A brute force approach would lead to the optimization of a sum of $K$ rational polynomials over a set of $|\setE|$ scalars. Instead, we proceed in a greedy manner, separately on each connected component of $\setE$, by starting with the mean consensus and optimizing successively subcrowns of increasing distances. All subcrowns of increasing distances are iteratively considered until $\lmsd(\tilde{T}, \setS)$ stops decreasing. For each subcrown $r = (\crown)^g$, we seek the scalar value $p_r\in[0,1]$ such that it minimizes

$$
p_r =\arg \min\limits_{x\in[0, 1]}({d}(\tilde{T}_{(td, g), x}, \setS))\text{, with } \tilde{T}_{(td, g), x} = \left\{
          \begin{array}{ll}
            x\text{ if }n\in r\\
            \tilde{T}_n\text{ otherwise}
          \end{array}
        \right.. 
\label{tacjacu_method}
$$
The algorithm is described in Alg.\ref{alg:TACJACU} and iteratively optimizes each subcrown from the inside to the outside of the $\setE$ set. We have observed no gain in combining a growing and a shrinking exploration of subcrowns contrary to Alg.~\ref{alg:hard}. 
For the optimization process of Eq.~\ref{tacjacu_method}, we use the SLSQP algorithm~\citep{SLSQP} implemented in Scipy v1.7.3~\citep{scipy}. Resulting consensus can be seen in Figs.~\ref{fig:4}, \ref{fig:6} and \ref{fig:7}.

\begin{algorithm}[!htpb]
\DontPrintSemicolon
\caption{Soft consensus algorithm}
\label{alg:TACJACU}
\SetKwInput{Initialization}{Initialization}
\SetKwComment{Comment}{ }{ }
\SetKw{and}{and}

\KwIn{$\setS\ \text{segmentation maps},\ \mathcal{N}\ \text{neighborhood},\ {d^s}\ \text{distance}$}
\KwResult{$\tilde{T}$}
\Initialization{Computation of $D_\setS^\mathcal{N};\ \tilde{T} = \frac{1}{K}\sum_{k=1}^K S^k$}
\While{$\lmsds(\tilde{T}, \setS)\text{ decreases}$}
{   
    \For{$td \in D_\setS^\mathcal{N}$ in increasing order}
    {   
        {   \For{$g \in \mathcal{P}(\llbracket 1, K \rrbracket)$}
            {   
                $p = \arg \min\limits_{x\in[0, 1]}(\lmsds(\tilde{T}_{(td, g), x}, \setS))$
                \Comment*[l]{with $\tilde{T}_{(td, g), x} = \left\{
                    \begin{array}{ll}
                        x\text{ on }(\crown)^g\\
                        \tilde{T}\text{ elsewhere }
                    \end{array}
                \right.$}
                $\tilde{T} \gets \tilde{T}_{(td, g), p}$
            }    

        }
    }
}
\end{algorithm}
 
\section{Results}

\subsection{Datasets and Implementation Details}
We tested our method on 3 datasets~: 
\begin{itemize}
    \item A private database of transition zones of prostate T2w MR images, composed of 40 cases segmented by 5 raters.
    \item The publicly available MICCAI MSSEG 2016 dataset of Multiple Sclerosis lesions segmentations~\citep{MSSEG} segmented from Brain MR images, with 15 subjects segmented by 7 raters
    \item The publicly available SCGM dataset~\citep{SCGM}, with 40 spinal cords and their grey matter segmented by 4 raters. We used the whole spinal cord segmentation (SCGM-SC) and the grey matter segmentation (SCGM-GM).
\end{itemize} Images from the private dataset (resp. MSSEG dataset, SCGM dataset) have a size of [80-288]$\times$[320-640]$\times$[320-640] voxels (resp. [144-261]$\times$[224-512]$\times$[224-512] voxels and [3-28]$\times$[100-655]$\times$[100-776] voxels). It was possible to extract from the private dataset bounding boxes of size [58-227]$\times$[53-184]$\times$[62-180] voxels. Similarly, we were able to extract from SCGM-SC (resp. SCGM-GM) bounding boxes of size [3-20]$\times$[15-90]$\times$[24-131] voxels(resp.)
From the 3D private dataset, we created a 2D subset by extracting a single slice for each patient located at the base of the prostate since this region is subject to a high inter-rater variability~\citep{BECKER,montagne_var}. \\
Examples for each dataset of segmentations by the different raters of the same case are available in Appendix~\ref{images_ir} (Fig.~\ref{fig:8}).

\paragraph{Implementation details} In the remainder, STAPLE results were produced by using the algorithm implemented in SimpleITK v2.0.2~\citep{simpleitk}. All MACCHIatO methods used the 8 or 26-connectivity neighborhood for 2D or 3D cases. MACCHIatO code is available at \href{https~://gitlab.inria.fr/dhamzaou/jaccardmap}{https~://gitlab.inria.fr/dhamzaou/jaccardmap}

\subsection{Heuristics relevance}
In Section~\ref{subcrown_def}, we have presented the subcrown-based heuristics that drives the optimization of the local mean square distance criteria. Indeed, those subcrown group voxels are based on three properties~: their morphological distance, the connected component they belong to, and the raters who segmented them. To check if this heuristics is appropriate, we compared it with two alternatives~: 
\begin{itemize}
  \item The first alternative iteratively minimizes the $\lmsd$ at the crown level {(as defined in subsection {\ref{subcrown_def}} and represented in Fig.~{\ref{fig:2a}})}, without any rater-related property.
  \item The other one iteratively processes each voxel separately.
\end{itemize}

We compared the 3 heuristics by computing a soft consensus (with the Tanimoto distance) on the toy example of Fig.~\ref{fig:5}, and we display their optimized value of $\lmsds$ and their computation time in Table~\ref{tab:3}. Furthermore, since the size of $\setE$ is small, we could estimate the true minimizer of $\lmsds$ that involves the optimization of $|\setE|$ parameters. 

Unlike the crown-based heuristics, the subcrown-based and voxel-based heuristics appear to compute a consensus close to the real $\lmsds$ minimizer. In addition, the subcrown method is significantly faster than the voxel-based approach.

We have also compared the three heuristics on two datasets in Table~\ref{tab:1}. The crown-based heuristics is the fastest method to compute but with the highest criteria $\lmsds$, whereas the voxel-based method requires far more computation time than the subcrown-based heuristics and even several hours for some Prostate 3D cases. Surprisingly, on average, the subcrown-based heuristics reaches a lower $\lmsds$ criteria than the voxel-based method, although the difference may hardly be seen in the produced consensus. On those datasets, we were not able to estimate the true minimizer of $\lmsds$, due to the high memory resources those computations would require.

\begin{table}
\begin{minipage}{0.45\textwidth}
\caption{Computed $\lmsds$ and computation time for the soft consensus with Tanimoto distance on the toy example of Fig.~\ref{fig:5} using three different heuristics and the true minimizer. }
\begin{tabular}{|c|c|c|}
  \hline
   Heuristics\ & $\lmsd$ & Time \\
   \hline
   \multirow{2}{*}{\shortstack{Subcrown-based\\heuristics}} & \multirow{2}{*}{\shortstack{0.159}} & \multirow{2}{*}{\shortstack{0.26s}}\\
   \ & \ & \ \\
   \hline
   \multirow{2}{*}{\shortstack{Crown-based\\heuristics}} & \multirow{2}{*}{\shortstack{0.176}} & \multirow{2}{*}{\shortstack{0.07s}}\\
   \ & \ & \ \\
   \hline
   \multirow{2}{*}{\shortstack{Voxel-based\\approach}} & \multirow{2}{*}{\shortstack{0.159}} & \multirow{2}{*}{\shortstack{0.92s}}\\
   \ & \ & \ \\
   \hline
   \multirow{2}{*}{\shortstack{Estimated True \\minimizer}} & \multirow{2}{*}{\shortstack{0.159}} & \multirow{2}{*}{\shortstack{0.55s}}
   \\
   \ & \ & \ \\
   \hline
\end{tabular}
\label{tab:3}

\end{minipage}
\hfill
\begin{minipage}{0.45\textwidth}
\includegraphics[width=0.75\textwidth]{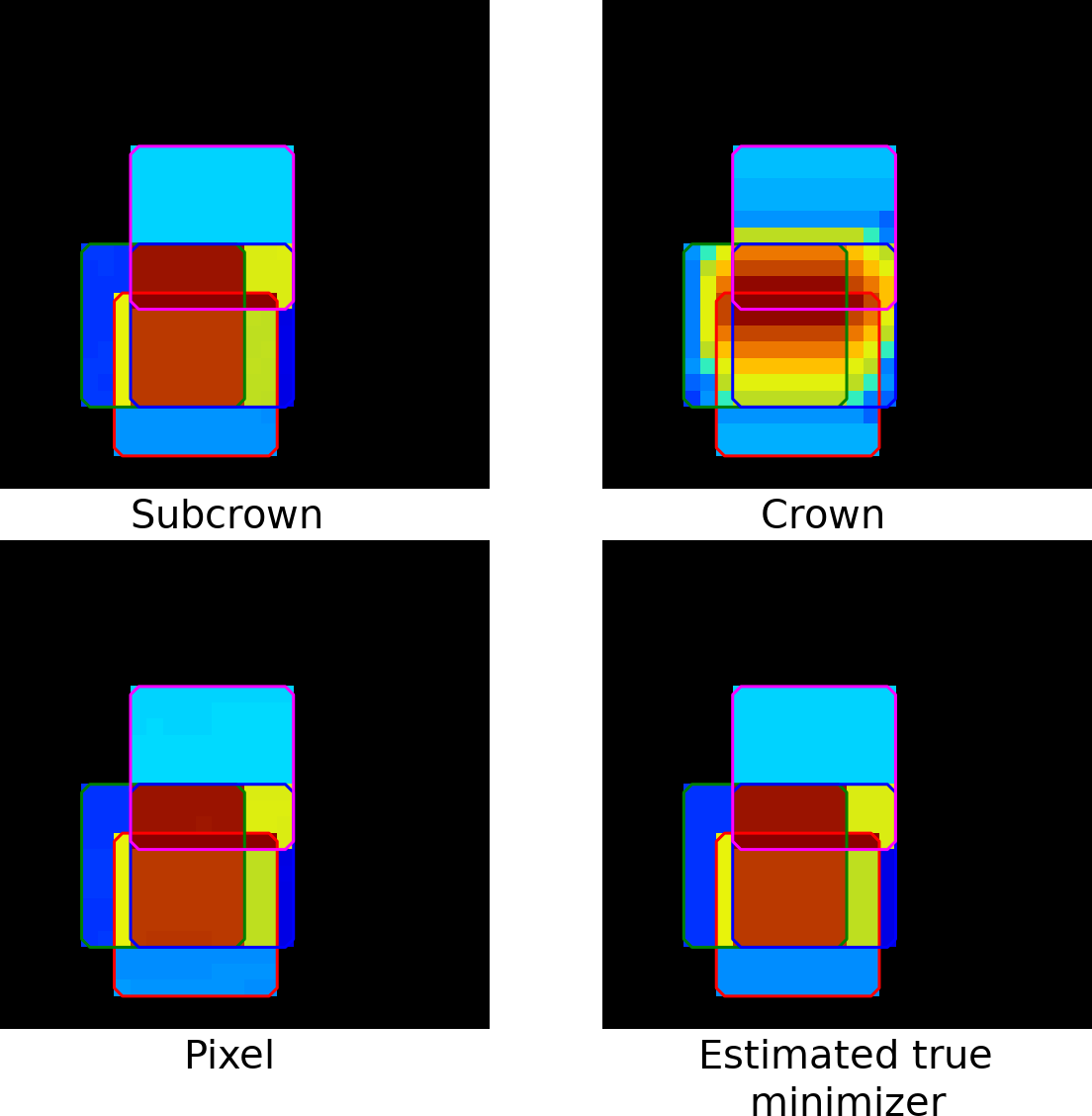}
\captionof{figure}{Different soft consensuses obtained on a toy example. Each contour corresponds to one of the raters' segmentation and colors indicate the probability using the same colormap as Fig~\ref{fig:4}.}
\label{fig:5}
\end{minipage}
\end{table}

\begin{table}[!htbp]
  \centering
  \caption{Mean $\lmsds$ and computation time for three different heuristics on some datasets}
  \begin{tabular}{|c|c|c|c|c|}
    \hline 
    Dataset & Subcrown & Crown & Voxel \\
    \hline
    MSSEG & 16.36 (57.48s) & 16.50 (23.41s) & 16.36 (20min30s)\\
    \hline
    Prostate 3D & 1.24e-2 (31.5s) & 1.26e-2 (5.46s) & NA \\
    \hline
    Prostate 2D & 5.98e-3 (0.29s) & 6.22e-3 (0.07s) & 6.10e-3 (5.30s)\\
    \hline
  \end{tabular}
  \label{tab:1}
\end{table}

\subsection{Comparison with baseline methods}


\paragraph{Comparison of inter-rater variabilities} A first set of experiments consist in measuring the impact of the choice of the consensus method when computing a measure of inter-rater variability. More precisely, we compute the average precision, recall, and F1-score between the hard consensus (considered as ground truth) and each rater segmentation. Those metrics have been computed on the MSSEG dataset where there are potentially large disagreements between raters. Table~\ref{tab:4_0} reports those metrics averaged among all lesions of all images, a lesion corresponding to a connected component of the mask union $\setE$.
The MV consensus has the highest recall and lowest precision which can be interpreted by a MV consensus smaller than other methods. Conversely, the STAPLE consensus has the largest precision
and lowest recall, thus corresponding to a larger size consensus. Regarding terms of F1-score, MV and MACCIHIatO methods obtained similar metrics but slightly higher for MACCHIatO-D (0.449).  
\begin{table}[!htbp]
    \centering
    \caption{Averaged lesion-wise measures on the MSSEG dataset for all hard consensus methods}
    \begin{tabular}{|c|c|c|c|c|}
    \hline
        \diagbox{Measure}{Method} & \shortstack{\ \\ML\\STAPLE} & MV & MACCHIatO-J & MACCHIatO-D\\ \hline
        Precision & 0.976 & 0.497 & 0.562 & 0.570\\ \hline
        Recall & 0.273 & 0.817 & 0.769 & 0.758 \\ \hline
        F1-score & 0.297 & 0.437 & 0.448 & 0.449\\ \hline
    \end{tabular}
    \label{tab:4_0}
\end{table}

In addition, we also compared the methods on the number of connected components. To do so, we defined each consensus as ground truth and from there computed the average precision, recall, and F1-score of each rater for lesion detection (considering the existence of a non-null intersection with the rater's segmentation as a sufficient threshold to detect). We performed this experiment on the MSSEG dataset, as it is our only dataset with several connected components per case. 
Table~\ref{tab:4} reports those metrics averaged among all patients.
The MV consensus has the highest detection recall and lowest detection precision which can be interpreted by a MV consensus not segmenting some lesions conserved by the other methods. Conversely, the STAPLE consensus has the largest precision
and lowest recall, thus corresponding to the presence of lesions rarely segmented by the raters. In terms of F1-score, MV and MACCHIatO methods are close to each other, but it is highest for MACCHIatO-D (0.894). 

\begin{table}[!htbp]
    \centering
    \caption{Measures of lesion detection on the MSSEG dataset for all hard consensus methods}
    \begin{tabular}{|c|c|c|c|c|}
    \hline
        \diagbox{Measure}{Method} & \shortstack{\ \\ML\\STAPLE} & MV & MACCHIatO-J & MACCHIatO-D\\ \hline
        Precision & 0.994 & 0.887 & 0.914 & 0.931\\ \hline
        Recall & 0.643 & 0.967 & 0.931 & 0.930\\ \hline
        F1-score & 0.746 & 0.892 & 0.888 & 0.894\\ \hline
    \end{tabular}
    \label{tab:4}
\end{table}

\begin{figure}
\centering
  \includegraphics[width=1.0\linewidth]{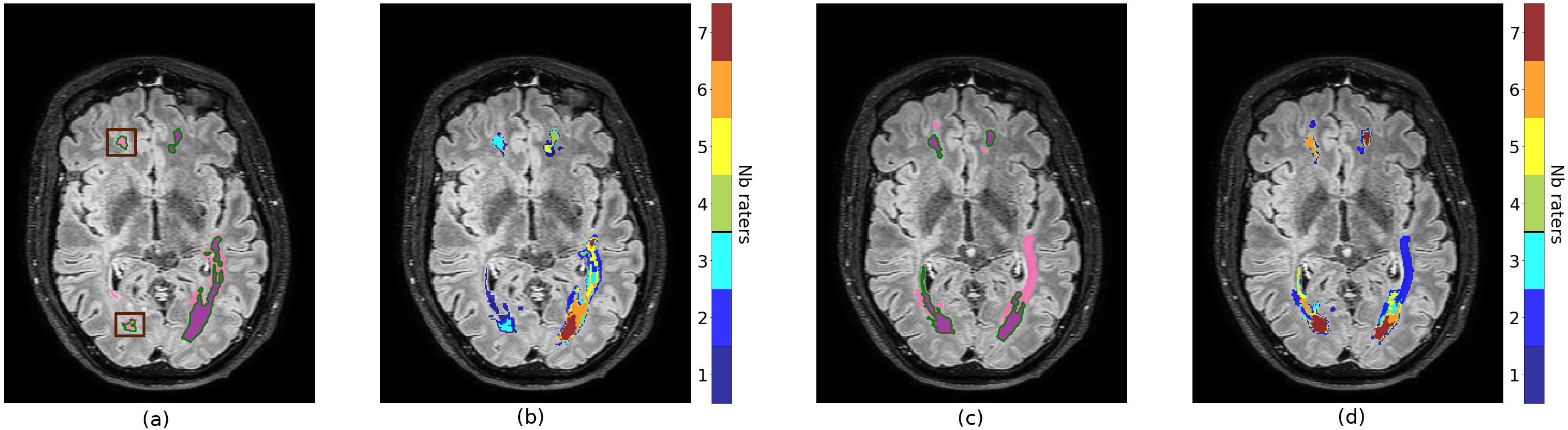}
\caption{Two consecutive slices of a MSSEG sample on which we applied STAPLE (pink), Majority Voting (purple) and MACCHIatO-TJ (green contour) (a, c), and for each voxel of those slices the number of raters who segmented them (b, d). We can note that some zones (highlighted by brown squares) were selected by soft MACCHIatO-TJ whereas less than the majority of raters segmented them.}
\label{fig:6}
\end{figure}

\paragraph{Comparison of consensus areas or volumes} In Table~\ref{tab:5}, we compare the relative size of hard consensuses on all datasets, taking the MV consensus as reference. On average, all methods lead to consensuses of larger size than MV. For the MACCHIatO methods, the difference with MV consensus is modest on a massive organ (prostate) but significant for small lesions ($>$16\%). The ML STAPLE method generates much larger consensuses than MV, especially when dealing with small lesions. Note that for the MSSEG dataset, ML STAPLE is computed on the whole image, thus with a large background size. Finally, the MACCHIatO-D and MACCHIatO-J methods lead to consensuses of similar size, without any clear order. Table~\ref{tab:6} compares the soft area or volumes of the soft consensuses (given by $\sum_{n=1}^N \tilde{U}_n$) generated by all methods, taking the mask averaging as reference. Fig.~\ref{fig:6} illustrates those soft consensuses on the MSSEG dataset. The variation of volumes is smaller for soft consensus than for hard consensus. In general, the MA method produces the smallest volumes, and STAPLE the largest ones. The methods using surrogate Dice or Jaccard distances give similar volumes, although the Soergel and $1SD$ are more diverging on the MSSEG dataset. We also compare the size of the thresholded maps $\tilde{U}_n>0.5$ which provide similar trends to their soft maps.




For both hard and soft consensuses, the largest differences between the different methods are observed on the MSSEG dataset, followed by SCGM-GM. 

\begin{table}[!htbp]
    \centering
    \caption{Left~: Average size variation on 3D datasets for hard consensuses, with the Majority Voting serving as the reference size. Right~: percentage of cases where the computed consensus is strictly larger than the MV consensus. Red color indicates that for this setting, all cases are at least of equal size.}
    \begin{tabular}{|c|c|c|c||c|c|c|}
        \cline{2-7}
        \multicolumn{1}{c}{\ } & \multicolumn{3}{|c||}{Avg. size variation w.r.t MV} & \multicolumn{3}{c|}{Frequencies of size $>$ $|\text{MV}|$}\\ \hline
        \diagbox{Dataset}{Method} & Jaccard & Dice & ML STAPLE & Jaccard & Dice & ML STAPLE\\ \hline
        Prostate 3D & \textcolor{red}{+0.4\%} & +0.6 \% & \textcolor{red}{+22\%} & \textcolor{red}{87.5\%} & 85\% & \textcolor{red}{100\%} \\ \hline
        MSSEG & \textcolor{red}{+19\%} & +16\% & \textcolor{red}{+151\%} & \textcolor{red}{100\%} & 93\% & \textcolor{red}{100\%} \\ \hline
        SCGM-SC & +2.36\% & +2.30\% & +11\% & 97.5\% & 97.5\% & \textcolor{red}{100\%}\\\hline
        SCGM-GM & \textcolor{red}{+17\%} & \textcolor{red}{+15\%} & \textcolor{red}{+47\%} & \textcolor{red}{100\%} & \textcolor{red}{100\%} & \textcolor{red}{100\%}\\\hline
    \end{tabular}
    \label{tab:5}
\end{table}

\begin{table}[!htbp]
    \centering
    \caption{Top~: Average soft volume variation on 3D datasets for soft consensuses, with the MA serving as the reference. Bottom~: Percentage of cases where the obtained consensus has a higher volume than the MA consensus. Red color indicates for the thresholded case that for this setting, all cases are at least of equal size.}
    \begin{tabular}{|c|c|c|c|c|c|}
        \cline{2-6}
        \multicolumn{1}{c}{\ } & \multicolumn{5}{|c|}{Avg. soft volume variation w.r.t MA}\\\hline
        \diagbox{Dataset}{Method} & TJ & SJ & 2SD & 1SD & STAPLE \\\hline
        Prostate 3D & +0.4\% & +0.1\% & +0.1\% & +0.7\% & +10\% \\ \hline
        Thresholded & \textcolor{red}{+0.1\%} & \textcolor{red}{+0.07\%} & +0.09\% & \textcolor{red}{+0.03\%} & \textcolor{red}{+11\%}\\ \hline\hline
        MSSEG & +4\% & \textcolor{red}{+16\%} & +2\% & -3\% & +43\%\\ \hline
        Thresholded & +8\% & \textcolor{red}{+37\%} & \textcolor{red}{+4\%} & \textcolor{red}{+11\%} & \textcolor{red}{+68\%}\\\hline\hline
        SCGM-SC & -0.4\% & +0.5\% & -0.5\% & +0.3\% & +4\% \\ \hline
        Thresholded & \textcolor{red}{+1\%} & \textcolor{red}{+1.3\%} & \textcolor{red}{+0.9\%} & \textcolor{red}{+0.9\%} & \textcolor{red}{+5.7\%} \\ \hline\hline
        SCGM-GM & +1.2\% & +4.4\% & +1\% & +2.9\% & +8.6\%\\ \hline
        Thresholded & \textcolor{red}{+13\%} & \textcolor{red}{+16\%} & \textcolor{red}{+11\%} & \textcolor{red}{+14\%} & \textcolor{red}{+19\%}\\\hline
    \end{tabular}
    \bigbreak
    \begin{tabular}{|c|c|c|c|c|}
         \cline{2-5}
         \multicolumn{1}{c}{\ } & \multicolumn{4}{|c|}{Frequencies of soft volume $>$ $|\text{MA}|$}\\ \hline
         \diagbox{Dataset}{Method} & TJ & SJ & 2SD & 1SD\\ \hline
         Prostate 3D & 80\% & 65\% & 60\% & 80\%\\ \hline
         Thresholded & \textcolor{red}{22.5\%} & \textcolor{red}{12.5\%} & 7.5\% & \textcolor{red}{7.5\%} \\ \hline\hline
         MSSEG & 87\% & \textcolor{red}{100\%} & 73\% & 33\%\\\hline
         Thresholded & 93\% & \textcolor{red}{100\%} & \textcolor{red}{80\%}  & \textcolor{red}{93\%}\\\hline\hline
         SCGM-SC & 10\% & 52.5\% & 5\% & 37.5\%\\ \hline
         Thresholded & \textcolor{red}{35\%} & \textcolor{red}{67.5\%} & \textcolor{red}{25\%} & \textcolor{red}{27.5\%}\\ \hline\hline
         SCGM-GM & 92.5\% & 95\% & 92.5\% & 82.5\%\\ \hline
         Thresholded & \textcolor{red}{100\%} & \textcolor{red}{100\%} & \textcolor{red}{100\%} & \textcolor{red}{100\%}\\\hline

    \end{tabular}

    \label{tab:6}
\end{table}

We recorded the cumulative running time for STAPLE and soft MACCHIatO methods to generate a consensus for all structures of our datasets in Table~\ref{tab:7}. We did not consider MA as it requires far less computation than the other methods. Among the considered algorithms STAPLE is in general the fastest method, being approximately 2-3 times faster than MACCHIatO methods. The exception here is the computation time on SCGM, which always involves small structure sizes and large image sizes. 

\begin{table}[htbp!]
    \centering
    \caption{Computation time of continuous methods on all datasets}

    \begin{tabular}{|c|c|c|c|c|c|c|}
        \hline
        \diagbox{Dataset}{Method} & TJ & SJ & 2SD & 1SD & STAPLE\\\hline
        Prostate 2D & 11.1s & 14.6s & 7.4s & 9.8s & 2.3s\\\hline
        Prostate 3D & 15m02s & 12m52s & 9m19s & 9m48s & 4m17s\\\hline
        MSSEG & 14m29s & 11m31s & 11m42s & 11m13s & 3m38s\\\hline
        SCGM-SC & 16.7s & 15.1s & 14s & 14.3 & 40.6s\\\hline
        SCGM-GM & 14.1s & 12.8s & 12.4s & 13.3s & 34.7s\\\hline
    \end{tabular}
    \label{tab:7}
\end{table}

\subsection{Entropy of soft consensus}

\begin{figure}[!htbp]
  \centering
  \includegraphics[width=1.0\linewidth]{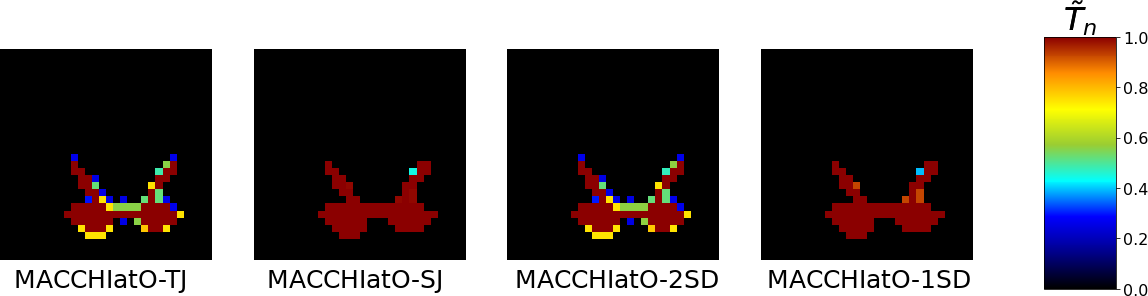}
  \caption{Impact of the choice of the distance on the computed soft MACCHIatO consensus on a SCGM-GM example}
  \label{fig:7}
\end{figure}

In Figs.~\ref{fig:3} and~\ref{fig:7} we show examples of soft consensuses on the prostate and grey matter datasets. It appears that MACCHIatO-SJ and MACCHIatO-1SD methods often assign to subcrowns probability values very close to 0 or 1 despite being soft consensus methods.
To confirm this behaviour, we compared on all 3D datasets the Shannon entropy $-\sum_n \tilde{U}_n\log \tilde{U}_n- (1-\tilde{U}_n)\log (1-\tilde{U}_n)$ obtained by MA and by the four soft MACCHIatO methods. Table~\ref{tab:8} confirms the strong binary behavior of MACCHIatO-SJ and MACCHIatO-1SD methods while MACCHIatO-TJ and MACCHIatO-2SD have a similar spread than mask averaging. Thus, we classify the surrogate distances between two families~: the ones associated with low-entropy consensus (Soergel, $d_{1SD}$), and the ones generating high-entropy consensus (Tanimoto, $d_{2SD}$). 
\begin{table}[htbp!]
  \centering
  \caption{Mean entropy on 3D datasets for soft MACCHIatO methods. MA entropy is given as a reference.}

  \begin{tabular}{|c|c|c|c|c|c|}
    \hline
     Dataset & MA & TJ & SJ & 2SD & 1SD  \\
     \hline
     Prostate 3D & 63850 & 63658 & 6928 & 63799 & 19361\\
     \hline
     MSSEG & 41295 & 37377 & 3805 & 37720  & 6107\\
     \hline
     SCGM-SC & 2401 & 2467 & 259 & 2483  & 305\\
     \hline
      SCGM-GM & 757 & 736 & 97 & 736 & 118\\
      \hline
  \end{tabular}
  \label{tab:8}
\end{table}

\subsection{Discussion}
 Experiments confirmed the dependence on background size of the STAPLE method, as shown in Fig.~\ref{fig:1a} and Appendix~\ref{STAPLE_back} (Tab.~\ref{tab:9}). 
 We also observed that hard consensuses obtained by MACCHIatO were generally slightly larger than those obtained by MV, particularly with MACCHIatO-J which almost never produces consensuses smaller than MV's. This can be explained by the fact that the MACCHIatO consensus may include voxels segmented by less than half of the raters (as seen in Figs.~\ref{fig:3} and~\ref{fig:6}). Finally, STAPLE consensuses always have a larger size than both MACCHIatO and MV. Similar observations can be made on soft consensus but with a smaller difference between methods on soft volumes compared to hard volumes. The MACCHIatO methods by construction create consensuses, independent from the background size, that maximize the local average (soft) squared Dice or Jaccard coefficients between the consensus and rater masks for each connected component. Furthermore, they produce masks that are different from the MV and STAPLE methods and have in general larger volumes than MV consensuses and smaller volumes than STAPLE ones. Finally, the MACCHIatO algorithms are in general more computationally expensive than MV or STAPLE algorithms but only to a reasonable extent (about 2 or 3 times more).  {In this article, we had the deliberate position not to choose between soft and hard consensus. From our perspective, the choice of method should be based on the users' motivations and the downstream task. If the users solely aim to generate a binary mask for visualization purposes or inter-rater variability studies, they can opt for the hard consensus method. However, if they wish to incorporate uncertainty modeling and obtain more refined results, the soft consensus methods would be more suitable.\\
Similarly, the choice of distances should align with the intended objectives. If users prioritize a solid mathematical foundation for the method, then they may opt for the Jaccard (hard) and Soergel (soft) metrics as, contrary to other used distances, they respect the triangular inequality. Alternatively, the Tanimoto distance can be used for uncertainty assessment, as MACCHIatO-TJ outputs more non-binary values than MACCHIatO-SJ. Users also have the flexibility to use the more commonly employed Dice instead of Jaccard if they prefer. In definitive, we have presented a range of methods within a consistent framework and elaborated on their characteristics. However, the specific configuration is ultimately left to the users based on their individual requirements and preferences}.
 
It can also be noted that the size variation observed on a dataset seems to be correlated with its inter-rater variability, the observed differences being more important on the MSSEG and SCGM-GM dataset than on the others.
 
In this article, we always considered 8-connexity in 2D cases and 26-connexity in 3D cases, as it performed better in preliminary experiments. However, the use of other neighborhoods (such as the 4-neighborhood in 2D, or the 6 and 18-neighborhood in 3D) could be envisaged. Moreover, we did not consider the case of highly anisotropic images, like in the SCGM dataset where a ratio of anisotropy greater than 10 in the voxel size is encountered. For those cases, it could be considered to apply a 2.5D approach consisting in applying our method to each slice independently. Comparisons between 2.5D and 3D neighborhoods on SCGM are available in Appendix~\ref{2.5D}.

The proposed method has several limitations. First, we only considered a binary segmentation problem. Extension to multiclass segmentation could be foreseen using for instance the generalization method presented in \cite{metric_extension} and \cite{soft_dice}. Second, the considered distances between binary sets are based on region overlap measures (Dice, Jaccard indices) and discard distances between boundaries such as Hausdorff Distance (HD). Our experiments based on HD were not conclusive. 

The reasons for this may be similar to the ones described in \cite{karimi_hd}~: instability of the methods to minimize a distance only defined from the largest error, HD sensitivity to outliers, difficulties to optimize it from an optimization point of view. To mitigate those effects, we made some tests using two of the Hausdorff alternatives defined in \cite{karimi_hd} and based respectively on distance maps and erosion, to no avail.\\
Third, the proposed criteria $\lmsd$, weights all raters equally for all connected components, unlike the STAPLE algorithm. It is possible to extend the MACCHIatO framework by attributing weights to raters based on their precision and recall (as those measures are independent of background size), either at the local or global level. Yet, this extension would require additional optimization steps, since the weights depend on the current estimate of the consensus.

Extending the MACCHIatO method to generate consensuses from $K$ (soft) probability maps instead of binary segmentations is not straightforward. Indeed, while minimizing the Fr\'{e}chet variance of Eq.~\ref{eq;variance-soft} is well-posed, we can no longer restrict its computation to the set $\setE$ and define subcrowns as optimization blocks. An alternative method that we have explored in our prior work\cite{audelan}, is to map probabilities to real values through a link function (e.g. a logit function) and then use robust parametric models (t-distributions) to fuse the probability maps.

\section{Conclusion}
In this paper, we have shown that the STAPLE method is impacted by the image background size and the choice of prior law. We have also introduced a new background-size independent method to generate a consensus based on Jaccard and Dice-based distances, thus extending the Majority Voting and mean consensus methods. More precisely, the generated masks minimize the average squared Jaccard or Dice distance between the consensus and each rater segmentation. The MACCHIatO algorithms are efficient and provide consistent masks by taking into account local morphological configurations between rater masks. The consensus masks are usually larger than those generated by the majority voting or mask averaging methods but smaller than those issued by STAPLE. 
Therefore, based on the experiments performed on three datasets, we believe that the hard and soft MACCHIatO algorithms are good alternatives to MV-based and STAPLE-based methods to define consensus segmentation.
\newpage

\acks{This work has been supported by the French government, through the 3IA Côte d’Azur Investments and UCA DS4H Investments in the Future project managed by the National Research Agency (ANR) with the reference numbers ANR-19-P3IA-0002 and ANR-17-EURE-0004 and by the Health Data Center of the AP-HP (Assistance Publique-Hôpitaux de Paris). 
Private data was extracted from the Clinical Data Warehouse of the Greater Paris University Hospitals (Assistance Publique-H\^opitaux de Paris).
We thank Julien Castelneau, software Engineer Inria, for his help in the development of MedInria Software (MedInria - Medical image visualization and processing software by Inria \url{https~://med. inria.fr/} - RRID~:SCR\_001462). The authors are grateful to the OPAL infrastructure from Université Côte d'Azur for providing resources and support. We also thank Alexandre Allera, Malek Ezziane, Anna Luzurier, Raphaelle Quint and Mehdi Kalai for providing prostate segmentations, Yann Fraboni and Etrit Haxholli for insightful discussions, and Federica Cruciani and Lucia Innocenti for feedback.

This paper is dedicated to the memory of our dear colleague Olivier Commowick who has been very active and innovative in the domain of data fusion.}

%
\ethics{The work follows appropriate ethical standards in conducting research and writing the manuscript, following all applicable laws and regulations regarding treatment of animals or human subjects.}

\coi{We declare we do not have conflicts of interest}

\bibliography{references}

\begin{thebibliography}{31}
\providecommand{\natexlab}[1]{#1}
\providecommand{\url}[1]{\texttt{#1}}
\expandafter\ifx\csname urlstyle\endcsname\relax
  \providecommand{\doi}[1]{doi: #1}\else
  \providecommand{\doi}{doi: \begingroup \urlstyle{rm}\Url}\fi

\bibitem[Aljabar et~al.(2009)Aljabar, Heckemann, Hammers, Hajnal, and
  Rueckert]{aljabar}
P.~Aljabar, R.A. Heckemann, A.~Hammers, J.V. Hajnal, and D.~Rueckert.
\newblock {Multi-atlas based segmentation of brain images: Atlas selection and
  its effect on accuracy}.
\newblock \emph{NeuroImage}, 46\penalty0 (3):\penalty0 726--738, 2009.
\newblock ISSN 1053-8119.
\newblock \doi{10.1016/j.neuroimage.2009.02.018}.

\bibitem[Asman and Landman(2012)]{asman}
Andrew Asman and Bennett Landman.
\newblock {Formulating Spatially Varying Performance in the Statistical Fusion
  Framework}.
\newblock \emph{Medical Imaging, IEEE Transactions on}, 31:\penalty0
  1326--1336, 06 2012.
\newblock \doi{10.1109/TMI.2012.2190992}.

\bibitem[Asman and Landman(2013)]{asman_2}
Andrew~J. Asman and Bennett~A. Landman.
\newblock {Non-local statistical label fusion for multi-atlas segmentation}.
\newblock \emph{Medical Image Analysis}, 17\penalty0 (2):\penalty0 194--208,
  2013.
\newblock ISSN 1361-8415.
\newblock \doi{10.1016/j.media.2012.10.002}.

\bibitem[Audelan et~al.(2020)Audelan, Hamzaoui, Montagne, Renard-Penna, and
  Delingette]{audelan}
Benoît Audelan, Dimitri Hamzaoui, Sarah Montagne, Raphaële Renard-Penna, and
  Hervé Delingette.
\newblock {Robust Fusion of Probability Maps}.
\newblock In Anne~L. Martel, Purang Abolmaesumi, Danail Stoyanov, Diana Mateus,
  Maria~A. Zuluaga, S.~Kevin Zhou, Daniel Racoceanu, and Leo Joskowicz,
  editors, \emph{Medical Image Computing and Computer Assisted Intervention -
  MICCAI 2020}, pages 259--268, Cham, 2020. Springer International Publishing.
\newblock ISBN 978-3-030-59719-1.

\bibitem[Becker et~al.(2019)Becker, Chaitanya, Schawkat, Muehlematter, Hötker,
  Konukoglu, and Donati]{BECKER}
Anton~S. Becker, Krishna Chaitanya, Khoschy Schawkat, Urs~J. Muehlematter,
  Andreas~M. Hötker, Ender Konukoglu, and Olivio~F. Donati.
\newblock Variability of manual segmentation of the prostate in axial
  t2-weighted mri: A multi-reader study.
\newblock \emph{European Journal of Radiology}, 121:\penalty0 108716, 2019.
\newblock ISSN 0720-048X.
\newblock \doi{https://doi.org/10.1016/j.ejrad.2019.108716}.

\bibitem[Commowick et~al.(2012)Commowick, Akhondi-Asl, and Warfield]{commowick}
Olivier Commowick, Alireza Akhondi-Asl, and Simon~K. Warfield.
\newblock {Estimating A Reference Standard Segmentation with Spatially Varying
  Performance Parameters: Local MAP STAPLE}.
\newblock \emph{{IEEE Transactions on Medical Imaging}}, 31\penalty0
  (8):\penalty0 1593--1606, August 2012.
\newblock \doi{10.1109/TMI.2012.2197406}.

\bibitem[Commowick et~al.(2018)Commowick, Istace, Kain, Laurent, Leray, Simon,
  Pop, Girard, Ameli, Ferr{\'e}, Kerbrat, Tourdias, Cervenansky, Glatard,
  Beaumont, Doyle, Forbes, Knight, Khademi, Mahbod, Wang, Mckinley, Wagner,
  Muschelli, Sweeney, Roura, Llado, Santos, Santos, Silva-Filho,
  Tomas-Fernandez, Urien, Bloch, Valverde, Cabezas, Vera-Olmos, Malpica,
  Guttmann, Vukusic, Edan, Dojat, Styner, Warfield, Cotton, and
  Barillot]{MSSEG}
Olivier Commowick, Audrey Istace, Michael Kain, Baptiste Laurent, Florent
  Leray, Mathieu Simon, Sorina~Camarasu Pop, Pascal Girard, Roxana Ameli,
  Jean-Christophe Ferr{\'e}, Anne Kerbrat, Thomas Tourdias, Fr{\'e}d{\'e}ric
  Cervenansky, Tristan Glatard, Jeremy Beaumont, Senan Doyle, Florence Forbes,
  Jesse Knight, April Khademi, Amirreza Mahbod, Chunliang Wang, Richard
  Mckinley, Franca Wagner, John Muschelli, Elizabeth Sweeney, Eloy Roura,
  Xavier Llado, Michel Santos, Wellington~P Santos, Abel~G Silva-Filho, Xavier
  Tomas-Fernandez, H{\'e}l{\`e}ne Urien, Isabelle Bloch, Sergi Valverde,
  Mariano Cabezas, Francisco~Javier Vera-Olmos, Norberto Malpica, Charles R~G
  Guttmann, Sandra Vukusic, Gilles Edan, Michel Dojat, Martin Styner, Simon~K.
  Warfield, Fran{\c c}ois Cotton, and Christian Barillot.
\newblock {Objective Evaluation of Multiple Sclerosis Lesion Segmentation using
  a Data Management and Processing Infrastructure}.
\newblock \emph{{Scientific Reports}}, 8\penalty0 (1):\penalty0 13650, December
  2018.
\newblock \doi{10.1038/s41598-018-31911-7}.

\bibitem[Crum et~al.(2006)Crum, Camara, and Hill]{metric_extension}
William Crum, Oscar Camara, and Derek Hill.
\newblock {Generalized Overlap Measures for Evaluation and Validation in
  Medical Image Analysis}.
\newblock \emph{IEEE transactions on medical imaging}, 25:\penalty0 1451--61,
  12 2006.
\newblock \doi{10.1109/TMI.2006.880587}.

\bibitem[Dai et~al.(2013)Dai, Ding, and Wahba]{exponentialFamily}
Bin Dai, Shilin Ding, and Grace Wahba.
\newblock {Multivariate Bernoulli distribution}.
\newblock \emph{Bernoulli}, 19\penalty0 (4):\penalty0 1465--1483, 2013.

\bibitem[Dewalle-Vignion et~al.(2015)Dewalle-Vignion, Betrouni, Baillet, and
  Vermandel]{staple_use}
Anne-Sophie Dewalle-Vignion, Nacim Betrouni, Clio Baillet, and Maximilien
  Vermandel.
\newblock {Is STAPLE algorithm confident to assess segmentation methods in PET
  imaging?}
\newblock \emph{Physics in Medicine and Biology}, 11 2015.
\newblock \doi{10.1088/0031-9155/60/24/9473}.

\bibitem[Deza and Deza(2016)]{deza}
Michel~Marie Deza and Elena Deza.
\newblock {Distances and Similarities in Data Analysis}.
\newblock In \emph{{Encyclopedia of Distances}}, pages 327--345, Berlin,
  Heidelberg, 2016. {Springer Berlin Heidelberg}.
\newblock ISBN 978-3-662-52844-0.
\newblock \doi{10.1007/978-3-662-52844-0_17}.

\bibitem[Gragera and Suppakitpaisarn(2018)]{diceinequality}
Alonso Gragera and Vorapong Suppakitpaisarn.
\newblock {Relaxed triangle inequality ratio of the Sørensen–Dice and
  Tversky indexes}.
\newblock \emph{Theoretical Computer Science}, 718:\penalty0 37--45, 2018.
\newblock ISSN 0304-3975.
\newblock \doi{https://doi.org/10.1016/j.tcs.2017.01.004}.
\newblock URL
  \url{https://www.sciencedirect.com/science/article/pii/S0304397517300361}.
\newblock WALCOM (Workshop on Algorithms and Computation).

\bibitem[Hamzaoui et~al.(2022)Hamzaoui, Montagne, Renard-Penna, Ayache, and
  Delingette]{mojito}
Dimitri Hamzaoui, Sarah Montagne, Rapha{\"e}le Renard-Penna, Nicholas Ayache,
  and Herv{\'e} Delingette.
\newblock Morphologically-aware jaccard-based iterative optimization (mojito)
  for consensus segmentation.
\newblock In Carole~H. Sudre, Christian~F. Baumgartner, Adrian Dalca, Chen Qin,
  Ryutaro Tanno, Koen Van~Leemput, and William~M. Wells~III, editors,
  \emph{Uncertainty for Safe Utilization of Machine Learning in Medical
  Imaging}, pages 3--13, Cham, 2022. Springer Nature Switzerland.
\newblock ISBN 978-3-031-16749-2.

\bibitem[Isensee et~al.(2021)Isensee, Jaeger, Kohl, Petersen, and
  Maier-Hein]{nnunet}
Fabian Isensee, Paul~F. Jaeger, Simon A.~A. Kohl, Jens Petersen, and Klaus~H.
  Maier-Hein.
\newblock {nnU-Net: a self-configuring method for deep learning-based
  biomedical image segmentation}.
\newblock \emph{Nature Methods}, 18:\penalty0 203--211, 2021.
\newblock \doi{10.1038/s41592-020-01008-z}.

\bibitem[Ji et~al.(2021)Ji, Yu, Wu, Ma, Bian, Bi, Li, Liu, Cheng, and
  Zheng]{arnaqueur}
Wei Ji, Shuang Yu, Junde Wu, Kai Ma, Cheng Bian, Qi~Bi, Jingjing Li, Hanruo
  Liu, Li~Cheng, and Yefeng Zheng.
\newblock Learning calibrated medical image segmentation via multi-rater
  agreement modeling.
\newblock In \emph{Proceedings of the IEEE/CVF Conference on Computer Vision
  and Pattern Recognition (CVPR)}, pages 12341--12351, June 2021.

\bibitem[Karimi and Salcudean(2019)]{karimi_hd}
Davood Karimi and Septimiu~E Salcudean.
\newblock {Reducing the Hausdorff Distance in Medical Image Segmentation with
  Convolutional Neural Networks}.
\newblock \emph{IEEE Transactions on medical imaging}, 39\penalty0
  (2):\penalty0 499--513, 2019.

\bibitem[Kosub(2019)]{kosub}
Sven Kosub.
\newblock {A note on the triangle inequality for the Jaccard distance}.
\newblock \emph{Pattern Recognition Letters}, 120:\penalty0 36--38, 2019.
\newblock ISSN 0167-8655.

\bibitem[Kraft(1988)]{SLSQP}
Dieter Kraft.
\newblock {A software package for sequential quadratic programming}.
\newblock Technical Report DFVLR-FB 88-28, DLR German Aerospace Center –
  Institute for Flight Mechanics, Koln, Germany, 1988.

\bibitem[Leach and Gillet(2007)]{leach2007}
Andrew~R. Leach and Valerie~J. Gillet.
\newblock {Similarity Methods}.
\newblock In \emph{{An Introduction To Chemoinformatics}}, pages 99--117,
  Dordrecht, 2007. Springer Netherlands.
\newblock ISBN 978-1-4020-6291-9.
\newblock \doi{10.1007/978-1-4020-6291-9_5}.

\bibitem[Lowekamp et~al.(2013)Lowekamp, Chen, Ibanez, and Blezek]{simpleitk}
Bradley Lowekamp, David Chen, Luis Ibanez, and Daniel Blezek.
\newblock {The Design of SimpleITK}.
\newblock \emph{Frontiers in Neuroinformatics}, 7, 2013.
\newblock \doi{10.3389/fninf.2013.00045}.

\bibitem[Ma et~al.(2021)Ma, Chen, Ng, Huang, Li, Li, Yang, and Martel]{junma}
Jun Ma, Jianan Chen, Matthew Ng, Rui Huang, Yu~Li, Chen Li, Xiaoping Yang, and
  Anne~L. Martel.
\newblock {Loss odyssey in medical image segmentation}.
\newblock \emph{Medical Image Analysis}, 71:\penalty0 102035, 2021.
\newblock ISSN 1361-8415.
\newblock \doi{https://doi.org/10.1016/j.media.2021.102035}.
\newblock URL
  \url{https://www.sciencedirect.com/science/article/pii/S1361841521000815}.

\bibitem[Montagne et~al.(2021)Montagne, Hamzaoui, Allera, Ezziane, Luzurier,
  Quint, Kalai, Ayache, Delingette, and Renard~Penna]{montagne_var}
Sarah Montagne, Dimitri Hamzaoui, Alexandre Allera, Malek Ezziane, Anna
  Luzurier, Raphaelle Quint, Mehdi Kalai, Nicholas Ayache, Herv{\'e}
  Delingette, and Raphaele Renard~Penna.
\newblock {Challenge of prostate MRI segmentation on T2-weighted images:
  inter-observer variability and impact of prostate morphology}.
\newblock \emph{{Insights into Imaging}}, 12\penalty0 (1), June 2021.
\newblock \doi{10.1186/s13244-021-01010-9}.

\bibitem[Prados et~al.(2017)Prados, Ashburner, Blaiotta, Brosch,
  Carballido-Gamio, Cardoso, Conrad, Datta, Dávid, Leener, Dupont, Freund,
  Wheeler-Kingshott, Grussu, Henry, Landman, Ljungberg, Lyttle, Ourselin,
  Papinutto, Saporito, Schlaeger, Smith, Summers, Tam, Yiannakas, Zhu, and
  Cohen-Adad]{SCGM}
Ferran Prados, John Ashburner, Claudia Blaiotta, Tom Brosch, Julio
  Carballido-Gamio, Manuel~Jorge Cardoso, Benjamin~N. Conrad, Esha Datta,
  Gergely Dávid, Benjamin~De Leener, Sara~M. Dupont, Patrick Freund, Claudia
  A.M.~Gandini Wheeler-Kingshott, Francesco Grussu, Roland Henry, Bennett~A.
  Landman, Emil Ljungberg, Bailey Lyttle, Sebastien Ourselin, Nico Papinutto,
  Salvatore Saporito, Regina Schlaeger, Seth~A. Smith, Paul Summers, Roger Tam,
  Marios~C. Yiannakas, Alyssa Zhu, and Julien Cohen-Adad.
\newblock {Spinal cord grey matter segmentation challenge}.
\newblock \emph{NeuroImage}, 152:\penalty0 312--329, 2017.
\newblock ISSN 1053-8119.
\newblock \doi{https://doi.org/10.1016/j.neuroimage.2017.03.010}.
\newblock URL
  \url{https://www.sciencedirect.com/science/article/pii/S1053811917302185}.

\bibitem[Rohlfing and Maurer(2007)]{rohlfing}
Torsten Rohlfing and Calvin~R. Maurer.
\newblock {Shape-Based Averaging}.
\newblock \emph{IEEE Transactions on Image Processing}, 16\penalty0
  (1):\penalty0 153--161, 2007.
\newblock \doi{10.1109/TIP.2006.884936}.

\bibitem[Sabuncu et~al.(2010)Sabuncu, Yeo, Van~Leemput, Fischl, and
  Golland]{5487420}
Mert~R. Sabuncu, B.~T.~Thomas Yeo, Koen Van~Leemput, Bruce Fischl, and Polina
  Golland.
\newblock A generative model for image segmentation based on label fusion.
\newblock \emph{IEEE Transactions on Medical Imaging}, 29\penalty0
  (10):\penalty0 1714--1729, 2010.
\newblock \doi{10.1109/TMI.2010.2050897}.

\bibitem[Späth(1981)]{spath}
H~Späth.
\newblock {The Minisum Location Problem for the Jaccard Metric}.
\newblock \emph{Operations-Research-Spektrum}, 3:\penalty0 91--94, 1981.

\bibitem[Sudre et~al.(2017)Sudre, Li, Vercauteren, Ourselin, and
  Jorge~Cardoso]{soft_dice}
Carole~H. Sudre, Wenqi Li, Tom Vercauteren, Sebastien Ourselin, and
  M.~Jorge~Cardoso.
\newblock {Generalised Dice Overlap as a Deep Learning Loss Function for Highly
  Unbalanced Segmentations}.
\newblock In M.~Jorge Cardoso, Tal Arbel, Gustavo Carneiro, Tanveer
  Syeda-Mahmood, Jo{\~a}o Manuel~R.S. Tavares, Mehdi Moradi, Andrew Bradley,
  Hayit Greenspan, Jo{\~a}o~Paulo Papa, Anant Madabhushi, Jacinto~C.
  Nascimento, Jaime~S. Cardoso, Vasileios Belagiannis, and Zhi Lu, editors,
  \emph{Deep Learning in Medical Image Analysis and Multimodal Learning for
  Clinical Decision Support}, pages 240--248, Cham, 2017. Springer
  International Publishing.

\bibitem[Virtanen et~al.(2020)Virtanen, Gommers, Oliphant, Haberland, Reddy,
  Cournapeau, Burovski, Peterson, Weckesser, Bright, {van der Walt}, Brett,
  Wilson, Millman, Mayorov, Nelson, Jones, Kern, Larson, Carey, Polat, Feng,
  Moore, {VanderPlas}, Laxalde, Perktold, Cimrman, Henriksen, Quintero, Harris,
  Archibald, Ribeiro, Pedregosa, {van Mulbregt}, and {SciPy 1.0
  Contributors}]{scipy}
Pauli Virtanen, Ralf Gommers, Travis~E. Oliphant, Matt Haberland, Tyler Reddy,
  David Cournapeau, Evgeni Burovski, Pearu Peterson, Warren Weckesser, Jonathan
  Bright, St{\'e}fan~J. {van der Walt}, Matthew Brett, Joshua Wilson, K.~Jarrod
  Millman, Nikolay Mayorov, Andrew R.~J. Nelson, Eric Jones, Robert Kern, Eric
  Larson, C~J Carey, {\.I}lhan Polat, Yu~Feng, Eric~W. Moore, Jake
  {VanderPlas}, Denis Laxalde, Josef Perktold, Robert Cimrman, Ian Henriksen,
  E.~A. Quintero, Charles~R. Harris, Anne~M. Archibald, Ant{\^o}nio~H. Ribeiro,
  Fabian Pedregosa, Paul {van Mulbregt}, and {SciPy 1.0 Contributors}.
\newblock {{SciPy} 1.0: Fundamental Algorithms for Scientific Computing in
  Python}.
\newblock \emph{Nature Methods}, 17:\penalty0 261--272, 2020.
\newblock \doi{10.1038/s41592-019-0686-2}.

\bibitem[Warfield et~al.(2004)Warfield, Zou, and Wells]{STAPLE}
S.K. Warfield, K.H. Zou, and W.M. Wells.
\newblock {Simultaneous truth and performance level estimation (STAPLE): an
  algorithm for the validation of image segmentation}.
\newblock \emph{IEEE Transactions on Medical Imaging}, 23\penalty0
  (7):\penalty0 903--921, 2004.
\newblock \doi{10.1109/TMI.2004.828354}.

\bibitem[Willett et~al.(1998)Willett, Barnard, and Downs]{willett}
Peter Willett, John~M. Barnard, and Geoffrey~M. Downs.
\newblock {Chemical Similarity Searching}.
\newblock \emph{Journal of Chemical Information and Computer Sciences},
  38\penalty0 (6):\penalty0 983--996, 1998.
\newblock \doi{10.1021/ci9800211}.

\bibitem[Zhang et~al.(2020)Zhang, Tanno, Xu, Jin, Jacob, Cicarrelli, Barkhof,
  and Alexander]{neurips_nn}
Le~Zhang, Ryutaro Tanno, Mou-Cheng Xu, Chen Jin, Joseph Jacob, Olga Cicarrelli,
  Frederik Barkhof, and Daniel Alexander.
\newblock {Disentangling Human Error from Ground Truth in Segmentation of
  Medical Images}.
\newblock In H.~Larochelle, M.~Ranzato, R.~Hadsell, M.F. Balcan, and H.~Lin,
  editors, \emph{Advances in Neural Information Processing Systems}, volume~33,
  pages 15750--15762. Curran Associates, Inc., 2020.

\end{thebibliography}


\clearpage
\appendix

\section{Influence of background size in STAPLE}

\label{STAPLE_back}
We can see that by definition $u_n$ is impacted by the value of $w_n$ and, through $TN_k$, by the background size $BS$ = $|\{n | \forall k, S_n^k = 0\}|$ (i.e. the number of voxels that no rater segmented). In the following subsections we will characterize the dependence of the produced consensus to those parameters.

\subsection{STAPLE dependence on background size at fixed foreground}
 
By definition, when the background size increases $TN_k$ also increases whereas $TP_k, FP_k$ and $FN_k$ remain constants. So, $ q_k \to 1$ when $BS \to \infty$ and we can write
\begin{align*}
  \logit{(u_n)} &\sim \logit{(w_n)} + \sum_{k, S_n^k=1} (\ln{(p_k)} - \ln{(1 - \frac{TN_k}{TN_k + FP_k})}) + \sum_{k, S_n^k=0} \ln{(1-p_k)}\\
  &\sim \logit{(w_n)} + \sum_{k, S_n^k=1} (\ln{(p_k)} - \ln{( \frac{FP_k}{N - B_k})}) + \sum_{k, S_n^k=0} \ln{(1-p_k)}\\
  &\sim \logit{(w_n)} + \sum_{k, S_n^k=1} (\ln{(N - B_k)} + \ln{(\frac{p_k}{FP_k})}) + \sum_{k, S_n^k=0} \ln{(1-p_k)}
\end{align*}

with $B_k$=$TP_k$+$FN_k$.

\subsection{Impact of the consensus prior \texorpdfstring{$w_n$}{} on the limit}
In \cite{STAPLE},
they proposed to set $w_n$ as a spatially uniform value $w_n=w$ where $w$ is either a constant (typically $w =0.5$) or defined as the average occurrence ratio ($w = \frac{1}{NK}\sum_{n,k} S_n^k$). We further consider more general priors of the form $w=\frac{A}{N^\alpha}$, with A a constant independent of the image size $BS$, thus having $\logit(w_n) = -\ln{(\frac{N^\alpha -A}{A})}$. 

From there, we can write
\begin{align*}
  \lim_{BS\xrightarrow{}\infty} \logit{(u_n)} &= -\ln{(\frac{N^\alpha-A}{A})} + \sum_{k, S_n^k=1} \ln{(N-B_k)} + \sum_{k, S_n^k=1} \ln{(\frac{p_k}{FP_k})} + \sum_{k, S_n^k=0} \ln{(1-p_k)}\\
  &= \sum_{k, S_n^k=1} \ln{(N-B_k)}-\ln{(N^\alpha-A)}+\ln{(A)} + \sum_{k, S_n^k=1} \ln{(\frac{p_k}{FP_k})} + \sum_{k, S_n^k=0}\ln{(1-p_k)}\\
  &\sim \sum_{k, S_n^k=1} \ln{(N)}-\alpha\ln{(N)}+\ln{(A)} + \sum_{k, S_n^k=1} \ln{(\frac{p_k}{FP_k})} + \sum_{k, S_n^k=0}\ln{(1-p_k)}
\end{align*}
And
\begin{align*}
  \lim_{BS\xrightarrow{}\infty} {u_n} &= \frac{1}{ 1 + \big{(}\frac{1}{A}\prod_k\frac{FP_k^{S_n^k}}{ p_k^{S_n^k} (1-p_k)^{1-S_n^k}}\big{)} N^{\alpha - \sum_k S_n^k}}
\end{align*}

\newpage
\begin{table}[H]
    \centering
    \caption{Mean soft consensus entropy and volume comparisons on Prostate 3D between STAPLE on the full image (Full size STAPLE) and on a bounded box surrounding the organ (Focused STAPLE). }
    \begin{tabular}{|c|c|c|c|}
        \hline
        Dataset & Measure & Full size STAPLE & Focused STAPLE \\\hline
        \multirow{2}{*}{Prostate 3D} & Entropy & 2019 & 10992 \\\cline{2-4}
        \ & Size & \ 300534 & 285329\ \\\hline\hline
        \multirow{2}{*}{SCGM-SC} & Entropy & 74 & 269 \\\cline{2-4}
        \ & Size & \ 11406 & 11275\ \\\hline\hline
        \multirow{2}{*}{SCGM-GM} & Entropy & 71 & 118 \\\cline{2-4}
        \ & Size & \ 1854 & 1838\ \\\hline
    \end{tabular}

    \label{tab:9}
\end{table}

\section{Proof of Majority Voting as a Fr\'{e}chet Mean}

\label{MV_proof}
With $S^1 , S^2, ..., S^K \in \{0, 1\}^N$ binary segmentation maps and $T$ their Fr\'{e}chet mean with regards to the function $\sqrt{A \triangle B} = \sqrt{|(A\cup B)\setminus(A\cap B)|}$, we have 
\begin{align*}
    T &= \arg \min_{M\in \{0, 1\}^N} \sum_{k} \big{(}\sqrt{|(S^k\cup M) \setminus (S^k\cap M)|}\big{)}^2 = \arg \min_{M\in \{0, 1\}^N} \sum_{k} \big{(}\sqrt{|(S^k\cup M) \setminus (S^k\cap M)|}\big{)}^2\\
    &= \arg \min_{M\in \{0, 1\}^N} \sum_{k}(\sum_{n} (S^k_n + M_n - S^k_n M_n)-S^k_n M_n) =\arg \min_{M\in \{0, 1\}^N} \sum_{k, n} {S^k_n}^2 + {M_n}^2 -2 S^k_n M_n\\
    &= \arg \min_{M\in \{0, 1\}^N} \sum_{n} \big{(}\sum_{k} (S^k_n - M_n)^2\big{)} =(\delta(\sum_{k} S^k_n > \frac{K}{2}))_n \text{(the Majority Voting consensus)}.
\end{align*}

\section{Inter-rater variability}
\label{images_ir}
\begin{figure}[!htbp]
  \centering
  \begin{subfigure}{0.7\linewidth}
  \includegraphics[width =1.0\linewidth, trim={0 0 20cm 0},clip]{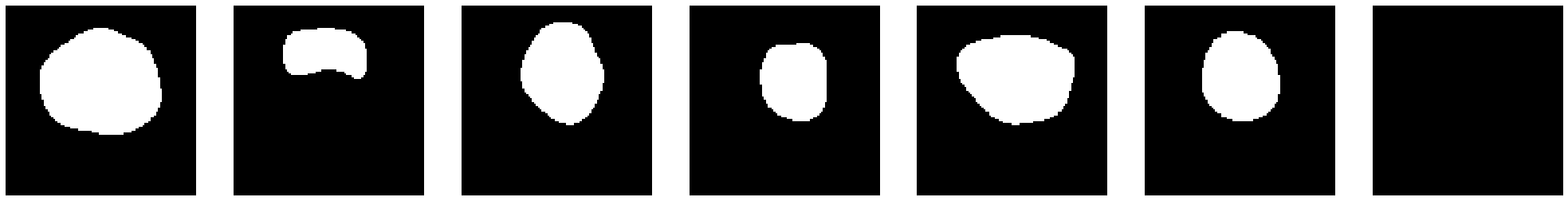}
  \subcaption{Prostate dataset}
  \label{fig:ST1a}
  \end{subfigure}
  
  \begin{subfigure}{0.7\linewidth}
  \includegraphics[width =1.0\linewidth]{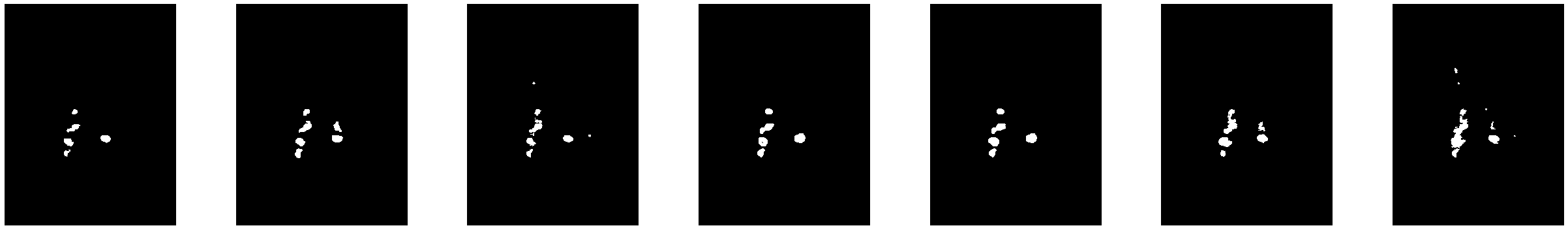}
  \subcaption{MSSEG}
  \label{fig:ST1b}
  \end{subfigure}
  
  \begin{subfigure}{0.45\linewidth}
  \centering
  \includegraphics[width =1.0\linewidth]{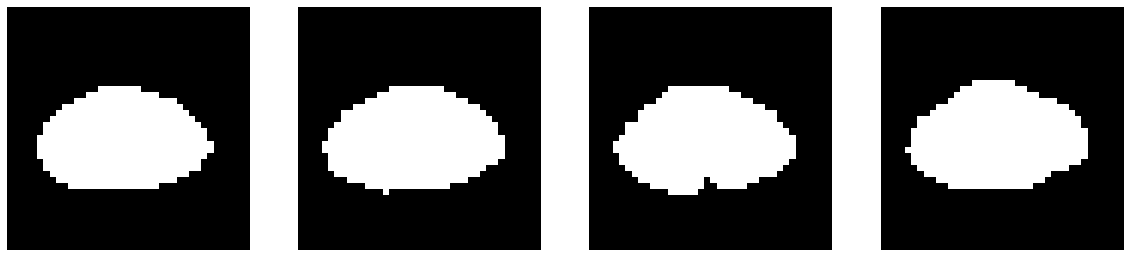}
  \subcaption{SCGM-SC}
  \label{fig:ST1c}
  \end{subfigure}
   \hfill
  \begin{subfigure}{0.45\linewidth}
  \centering
  \includegraphics[width =1.0\linewidth]{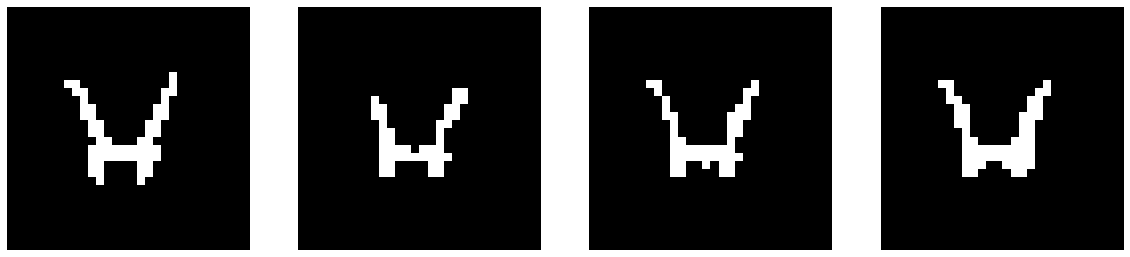}
  \subcaption{SCGM-GM}
  \label{fig:ST1d}
  \end{subfigure}
  
  \caption{Example of the inter-rater variability between the raters for the different datasets.}
  \label{fig:8}
\end{figure}

\section{Comparison between 2.5D and 3D neighborhoods}
\label{2.5D}

\begin{table}[!htbp]
\centering
\caption{Size comparisons for hard MACCHIatOs between the 2.5D and 3D neighborhood on SCGM-SC (top) and SCGM-GM (bottom)}
\begin{minipage}{1.0\linewidth}
    \begin{tabular}{|c|c|c|c|c|}
        \cline{2-5}
        \multicolumn{1}{c}{\ } & \multicolumn{2}{|c|}{Avg. size variation w.r.t MV} & \multicolumn{2}{|c|}{Direct size comparisons}\\\hline
        Method & 3D & 2.5D & $|\text{3D}|>|\text{2.5D}|$ & $|\text{3D}|<|\text{2.5D}|$    \\ \hline
        Jaccard & +2.37\% & +1.66\% & 32.5\% & 65\% \\ \hline
        Dice & +2.3\% & +1.6\% & 37.5\% &  55\% \\ \hline
    \end{tabular}
    \subcaption{SCGM-SC}
\end{minipage}
    \begin{minipage}{1.0\linewidth}
    \begin{tabular}{|c|c|c|c|c|}
        \cline{2-5}
        \multicolumn{1}{c}{\ } & \multicolumn{2}{|c|}{Avg. size variation w.r.t MV} & \multicolumn{2}{|c|}{Direct size comparisons}\\\hline
        Method & 3D & 2.5D & $|\text{3D}|>|\text{2.5D}|$ & $|\text{3D}|<|\text{2.5D}|$   \\ \hline
        Jaccard & +16.9\% & +15.8\% & 77.5\% & 15\% \\ \hline
        Dice & +14.7\% & +14.9\% & 67.5\% & 27.5\% \\ \hline
    \end{tabular}
    \subcaption{SCGM-GM}
    \end{minipage}

    \label{tab:10}
\end{table}

\begin{figure}[!htbp]
    \raggedright
    \includegraphics[width = 0.8\linewidth]{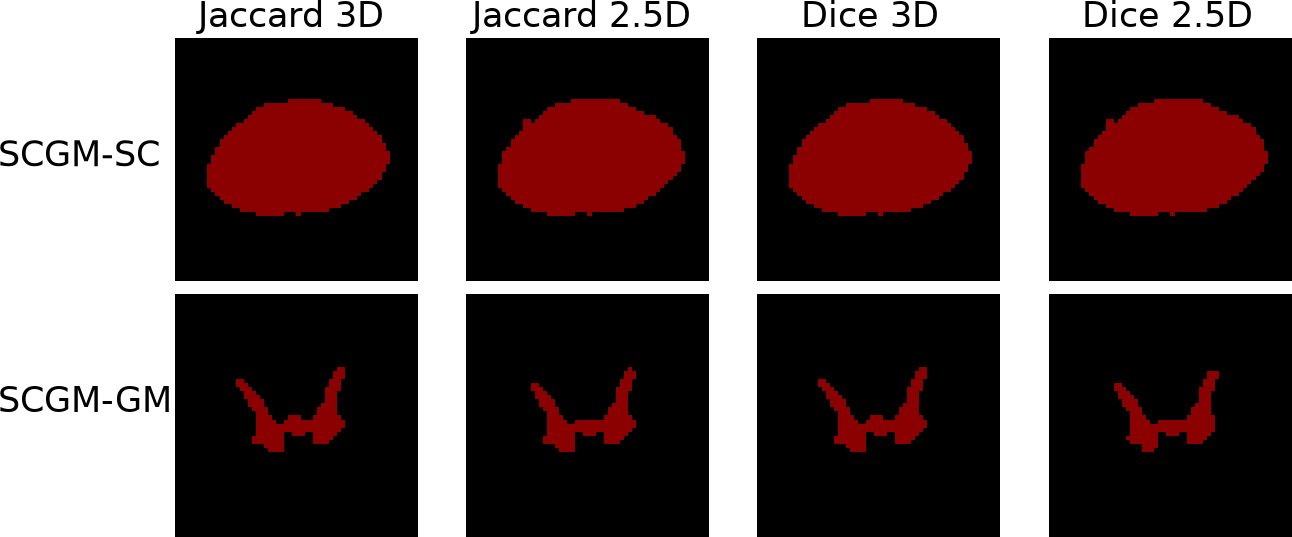}
    \caption{Examples of hard consensuses on SCGM with 2.5D and 3D neighborhoods.}
    \label{fig:final}
\end{figure}

\

\end{document}